\documentclass{article}

\usepackage[final]{neurips_2023}

\usepackage[utf8]{inputenc} %
\usepackage[T1]{fontenc}    %
\usepackage{hyperref}       %
\usepackage{url}            %
\usepackage{booktabs}       %
\usepackage{amsfonts}       %
\usepackage{nicefrac}       %
\usepackage{microtype}      %
\usepackage[table]{xcolor} %
\usepackage{wrapfig}

\usepackage{bm}

\def\eqref#1{equation~\ref{#1}}

\def\1{\bm{1}}

\def\vt{{\bm{t}}}

\def\vx{{\bm{x}}}
\def\vy{{\bm{y}}}
\def\vz{{\bm{z}}}

\def\vt{{\bm{t}}}

\def\vx{{\bm{x}}}
\def\vy{{\bm{y}}}
\def\vz{{\bm{z}}}

\def\mF{{\bm{F}}}

\DeclareMathAlphabet{\mathsfit}{\encodingdefault}{\sfdefault}{m}{sl}
\SetMathAlphabet{\mathsfit}{bold}{\encodingdefault}{\sfdefault}{bx}{n}

\usepackage{xspace}
\newcommand{\enc}{\mathcal{E}}
\newcommand{\dec}{\mathcal{D}}
\newcommand{\ours}{\textsc{PLANNER}\xspace}
\newcommand{\ie}{{\it i.e.}\xspace}
\newcommand{\textred}[1]{\textcolor{red}{#1}}

\usepackage{tabulary}
\usepackage{amsmath}
\usepackage{amssymb}
\usepackage{MnSymbol}%
\usepackage{wasysym}%
\usepackage{tikz}
\usepackage{xcolor}
\usepackage{color, colortbl}
\usepackage[normalem]{ulem}
\usepackage{url}
\usepackage{multirow}
\usepackage{float}
\usepackage{paralist}
\usepackage{comment}

\usepackage{mathtools}
\usepackage{arydshln}
\usepackage{subcaption}
\usepackage{enumitem}

\usepackage[textwidth=0.8in,textsize=tiny]{todonotes}
\usepackage{booktabs}
\usepackage{soul}

\usepackage{array}
\newcolumntype{H}{>{\setbox0=\hbox\bgroup}c<{\egroup}@{}}

\setlist{noitemsep,topsep=0pt,parsep=0pt,partopsep=0pt, leftmargin=12pt}

\usepackage[capitalize]{cleveref}
\crefformat{equation}{Eq.~(#2#1#3)}
\crefname{section}{§}{§§}
\Crefname{section}{§}{§§}

\title{\ours: Generating Diversified Paragraph via Latent Language Diffusion Model}

\author{
Yizhe Zhang, Jiatao Gu, Zhuofeng Wu, Shuangfei Zhai, Josh Susskind, Navdeep Jaitly \\
Apple\\
\texttt{\{yizzhang, jgu32, zhuofeng\_wu, szhai, jsusskind, njaitly\}@apple.com}
}

\begin{document}

\maketitle

\begin{abstract}

Autoregressive models for text sometimes generate repetitive and low-quality output because errors accumulate during the steps of generation. This issue is often attributed to exposure bias -- the difference between how a model is trained, and how it is used during inference. 
Denoising diffusion models provide an alternative approach in which a model can revisit and revise its output. However, they can be computationally expensive and prior efforts on text have led to models that produce less fluent output compared to autoregressive models, especially for longer text and paragraphs. In this paper, we propose \ours, a model that combines latent semantic diffusion with autoregressive generation, to generate fluent text while exercising global control over paragraphs. 
The model achieves this by combining an autoregressive ``decoding'' module with a ``planning'' module  that uses latent diffusion to generate semantic paragraph embeddings in a coarse-to-fine manner. 
The proposed method is evaluated on various conditional generation tasks, and results on semantic generation, text completion and summarization show its effectiveness in generating high-quality long-form text in an efficient manner.

\end{abstract}

\section{Introduction}

\begin{figure}[ht!]
    \centering
    \includegraphics[width=1.0\linewidth]{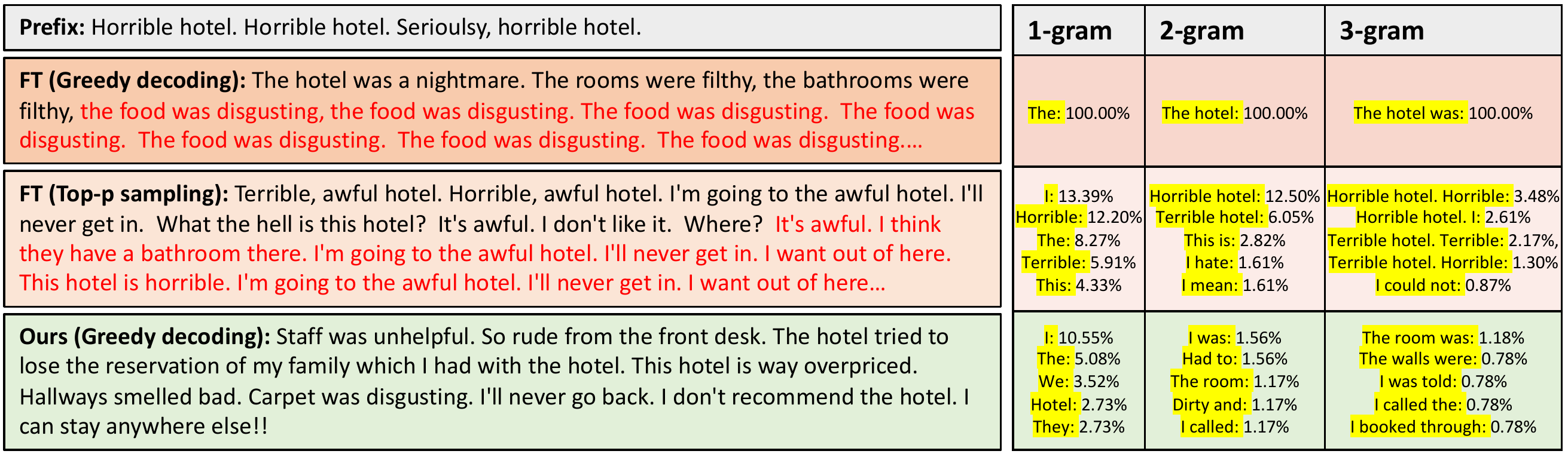}
    \caption{Left: With a repetitive prompt, the finetuned GPT-2 large model (774M, \textbf{FT}) is still attracted to 
    self-reinforced repetition (\textred{highlighted text}) even under top-p sampling (K=50, p=0.92). Right: the most frequent first n-grams of the generations for each method, derived from 512 generation roll-outs. Our proposed method results in a more diversified generation robust to the ill-composed prompt.
    }
    \label{fig:repetition}
\end{figure}

Autoregressive models trained with a teacher forcing strategy \citep{williams1989learning} are considered the gold standard for text generation. However, a significant drawback of this approach is that it lacks the ability to correct any mistakes made during the generation process which can lead to errors that accumulate as the generation progresses. 
Previous work \citep{ott2018analyzing,holtzmancurious,welleck2019neural,xulearning} has observed that \textit{deterministic decoding} methods have a tendency to generate consecutive repetitions at the word, phrase and sentence levels.
For example, with repetitive prompt, the model can enter an \textit{absorbing state} where it produces repetitive outputs with higher and higher confidence by \textit{self-reinforcing} the pattern \citep{xulearning}        (Fig.~\ref{fig:repetition}). Through our experiments, we have observed that such degeneration is more prevalent in \textit{open-ended} tasks that allow the model greater freedom for creativity.
Even for large language models, the generation can drift away from the desired semantics, especially when the model is poorly prompted or has high initial probabilities \citep{xulearning}.

Why does using maximum likelihood decoding lead to repetitions during generation which is significantly different from the training data distribution? 
One possible explanation for this is ``exposure bias'' \citep{bengio2015scheduled} arising from the discrepancy between the training and inference phases in the teacher forcing training strategy. During training phase, the model focuses only on predicting the next token. However, during inference, predicting the next token alone can be myopic because the model may not have enough foresight to anticipate its impact on future generation. %
This can also be seen as the ``distribution shift'' issue of behavior cloning \citep{imitation_learning}, where the model is trained to mimic the expert's actions on the states encountered by the expert in the training data. However, small differences between the model and the expert can compound over multiple steps, leading the model to states it never encountered during training, rendering  unreliable and undesirable predictions.

Although many approaches have been proposed to address this issue, such as adversarial models \citep{yu2017seqgan, lamb2016professor, zhang2017adversarial},  reinforcement learning \citep{li2016deep} or repetition penalties \citep{xulearning}
, they attempt to improve the \textit{global} aspects of the generation by making \textit{local} adjustments which still follow the autoregressive generation recipe. 
Diffusion models provide an alternative solution -- the model can revisit and revise its output iteratively, potentially rendering more global control of the generation in a non-autoregressive manner. However, these text diffusion models can generate less fluent text compared to autoregressive ones \citep{gong2022diffuseq}. Also, when generating long text, the diffusion process involves multiple passes of the underlying denoising model over a long generation length, making it computationally expensive. The discrete nature of text also presents a challenge for diffusion models, which can suffer from ``rounding errors'' when converting between the text token and its embedding \citep{li2022diffusion, lin2022genie}.

Instead of performing diffusion on the original text or the corresponding word embeddings, we propose to apply diffusion techniques to the latent semantic space \citep{rombach2022high, lovelace2022latent}. To achieve this, we learn a fixed number of continuous semantic tokens that encode salient information at the paragraph level. These tokens can then be used to reconstruct the original text. The latent diffusion can be additionally conditioned on an external signal to generate the semantic tokens. Finally, a decoder maps the obtained semantic tokens back to the raw text space. This process combines a non-autoregressive semantic diffusion approach with an autoregressive decoding technique.
The semantic diffusion process handles the ``planning'', enabling the modification of semantics in a coarse-to-fine manner, while the decoder handles the ``decoding'' by translating the semantics into raw text, with less flexibility in controlling the meaning. 
We call our proposed method \ours (\textbf{P}aragraph-leve\textbf{L} Diffusio\textbf{N} model for \textbf{E}mbedding \textbf{R}epresentation)\footnote{Our code is available at \url{https://github.com/apple/ml-planner}}.

Our contributions include:
$(i)$ We propose a latent semantic diffusion model for paragraphs that incorporates both non-autoregressive semantic diffusion and autoregressive generation. This allows us to generate fluent text while being able to exercise global control inherited from a diffusion model.
$(ii)$ We study the essential requirements for a good latent space for paragraph diffusion models.
$(iii)$ We evaluate the effectiveness of our proposed method on various conditional generation tasks.
Thanks to the iterative refinement of desnoising diffusion, our method enjoys less repetitive and more diverse generation, while maintaining good fluency and relevancy, comparing with autoregressive and text diffusion baselines \citep{li2022diffusion, lin2022genie}.

\section{Preliminary}
\paragraph{Diffusion Probabilistic Models}
The standard diffusion model (DM) \citep{ho2020denoising, song2019generative} learns the data distribution $p(x)$ by gradually denoising a normally distributed variable in a Markov chain of length $T$. The diffusion process can be viewed as a continuous-time stochastic process \citep{song2020score,kingma2021variational} where the initial data point $\vx \in \mathbb{R}^N$ is progressively corrupted by noise according to a predefined signal-noise schedule $\{\alpha_t, \sigma_t\}$, resulting in time-dependent corrupted data $\{\vx_t | t\in [0, 1], \vx_0=\vx\}$. The transition distribution
is given by: 
\begin{align}
    q(\vx_t | \vx_s) = \mathcal{N}(\vx_t; \alpha_{t|s}\vx_s, \sigma^2_{t|s}I), 
\end{align}
where $\alpha_{t|s} = \alpha_t/\alpha_s, \sigma^2_{t|s}=\sigma_t^2-\alpha_{t|s}^2\sigma_s^2$, and $s < t$. 
When $\vx_s=\vx$, the marginal distribution $q(\vx_t|\vx)$ is given as $q(\vx_t | \vx) = \mathcal{N}(\vx_t;\alpha_t\vx, \sigma_t^2I)$. 
The diffusion model relies on a parametric function $\theta$ optimized to reverse the diffusion process by denoising $\vx_t$ to the clean input $\vx$. The model is trained using a weighted reconstruction loss: 
\begin{align}
    \mathcal{L}(\theta) = \mathbb{E}_{\vx_t\sim q(\vx_t | \vx), t \sim [0, 1]}\left[\omega_t\cdot\|\mF_\theta(\vx_t, t) - \vx\|_2^2\right],
\end{align}
where $\omega_t = {\alpha_t^2}/{\sigma_t^2}, (s.t. \ \alpha_t^2 + \sigma_t^2 = 1$) is the signal-to-noise-ratio (SNR) and $\mF_\theta(\cdot)$ denotes the backbone denoising function. 
Sampling from the learned model can be performed using either ancestral sampling (DDPM) \citep{ho2020denoising} or a deterministic DDIM sampler \citep{song2021denoising}. 
While the DM is capable of generating high-quality samples, the fact that the corrupted data $\vx_t$ shares the same space as the input $\vx$ results in inefficient training \citep{jing2022subspace} and difficulty in learning abstract and semantically meaningful latent spaces \citep{preechakul2022diffusion}.

\paragraph{Latent Diffusion Models}
To improve the efficiency, the Latent Diffusion Model (LDM) \citep{rombach2022high} introduces an explicit separation between the compressive and generative learning phases of training diffusion models. It employs an autoencoding model consisting of an encoder $\enc(\cdot)$ and a decoder $\dec(\cdot)$ to learn a low-dimensional latent space that is perceptually equivalent to the image space when decoded, but with reduced computational complexity, while retaining the perceptual quality of generated samples. The reweighted objective for training LDM is given by:
\begin{align}
    \mathcal{L}(\theta) = \mathbb{E}_{\vz_t\sim q(\vz_t | \vz), \vz = \enc(\vx), t \sim [0, 1]}\left[\omega_t\cdot\|\mF_\theta(\vz_t, t) - \vz\|_2^2\right],
\end{align}
where $\vz$ is obtained from $\enc$ during training. The generated $\vz$ can be decoded to image using $\dec$.

\section{Related Work}
\paragraph{Text diffusion models} 
Early attempts on using diffusion models for discrete data used a noising processes which masked or randomly mutated the discrete tokens \citep{austin2021structured, hoogeboomautoregressive}. Recently, Diff-LM \citep{li2022diffusion} and DiffuSeq \citep{gong2022diffuseq} have instead used a continuous token embedding space, converting the continuous token embeddings to text via "rounding". Analog Bits \citep{chen2022analog} converts raw text into a set of bits and models them as analog bits with a continuous diffusion model. \citep{lovelace2022latent} performed diffusion model on the contextualized BART embeddings rather than on the word embedding space. \citep{zhu2022exploring} has applied text diffusion to image-captioning and achieved good performance. 

However, existing text diffusion models present several issues:
\textit{(i)} The varying length of the input text necessitates the prediction of additional length or superfluous paddings,
and \textit{(ii)} token generation in parallel may result in disfluent text and/or frequent repetitions especially when the generation is long. 
We instead employ the diffusion model to learn paragraph embeddings that contain 
fewer fixed-sized tokens, which allows for computational benefits and improved fluency.

\paragraph{Text Variational Autoencoders} 
Text VAEs \citep{bowman2016generating, kim2018tutorial, li2020optimus} have been particularly useful for learning a smooth and interpretable representation space, as well as for generating diverse text. However, one of the challenges is the KL vanishing problem \citep{bowman2016generating}, which results in the decoder disregarding the latent code sampled from the prior distribution during the inference stage. 
Our approach can be perceived as to address this issue by leveraging a more flexible prior distribution to ensure the codes can strongly influence the output text distribution.

\section{\ours: A Language Diffusion Model on Paragraph Embeddings}
\begin{figure}[ht!]
    \centering
    \includegraphics[width=1.0\linewidth]{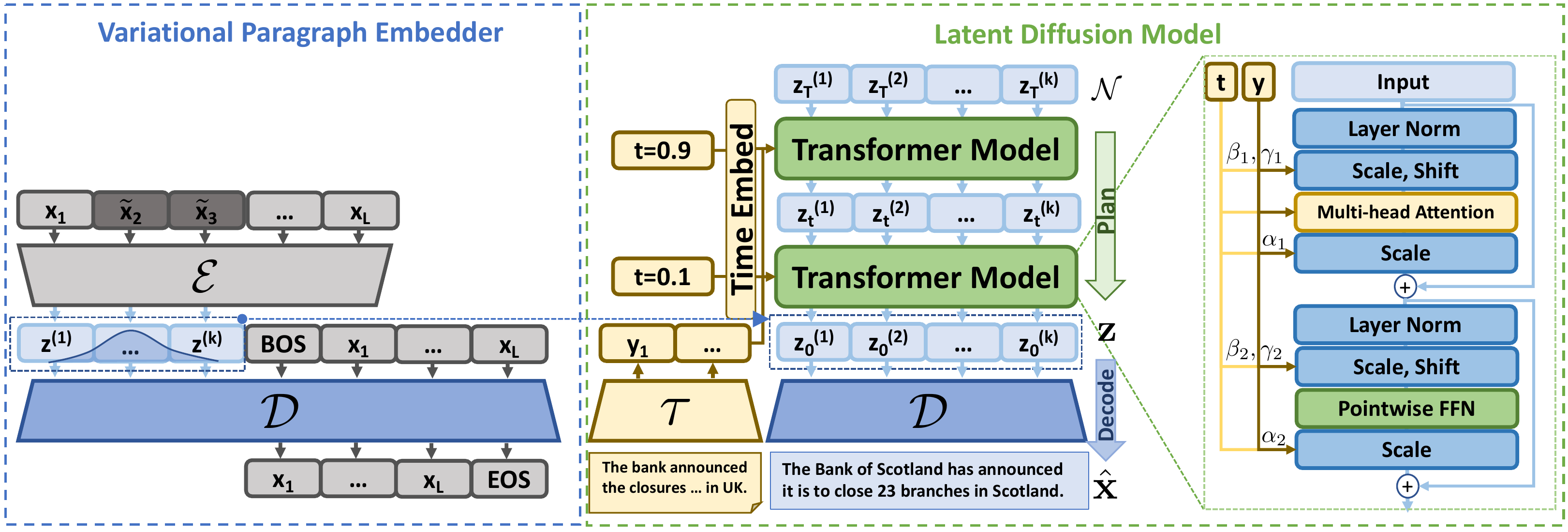}
    \caption{Model overview. Left: a variational paragraph embedder is learned to encode paragraph into a fixed amount of latent codes. Right: the latent diffusion model based on transformer block is applied to generate the latent codes. The decoder finally translates them into the text. (\textit{BOS}: Begin of Sentence token, \textit{EOS}: End of Sentence token)}
    \label{fig:model}
\end{figure}

We use latent diffusion to improve the diversity and fluency of paragraphs generated from the model. 
Our model comprises two parts (Fig.~\ref{fig:model}) - a paragraph embedder via variational autoencoder (VAE) that learns a meaningful and smooth latent space that corresponds to the original text space, and a diffusion model that generates latent codes corresponding to the semantics of longer text.

\subsection{Learning a Variational Paragraph Embedder}
Instead of applying diffusion to tokens directly to generate long text, we propose to learn a set of latent codes $\vz = \{z^{(1)},\cdots,z^{(k)}\} \in \mathbb{R}^{k \times h}$, which we call \textit{paragraph embeddings}, that capture the semantics in the target text (of length up to 512 tokens), where $h$ denotes the embedding dimension. These paragraph embeddings have shorter length, such as $k=16$, than the original text.

To obtain such embeddings $\vz$, we train a transformer-based encoder-decoder model. The architecture used for the autoencoder is shown in Fig.~\ref{fig:model}.
The encoder $\enc$ and decoder $\dec$ construct a bidirectional mapping between the discrete data space and the latent code space.  The paragraph embeddings $\vz$ are extracted by taking the first $k$ hidden state vectors of dimension $h$ from the final layer of $\enc$, which are fed into the initial steps of the decoder which is trained to reconstruct the original text. It's worth noting that the paragraph embeddings share the same hidden dimension $h$ as the word embeddings, and forming a manifold in the word embedding space. 
Pretrained BERT and GPT-2 models are used to initialize $\enc$ and $\dec$, respectively. 
The manifold of the learned  embeddings ideally possesses several desirable properties, including low \textit{conversion error}, \textit{local smoothness} and \textit{distributional smoothness}.

\paragraph{Conversion error} Ideally, the original input $\vx$ can be perfectly reconstructed via $\hat{\vx}= \dec(\vz), \vz=\enc(\vx)$, and modeling the lower-dimensional continuous space $p(\vz)$ is equivalent to modeling $p(\vx)$. However, in practice a loss of information can occur when converting raw text into paragraph embeddings or when doing the reverse.
We assess the conversion loss by computing the BLEU score ($\textbf{BLEU}_{\texttt{clean}}$) between the input $\vx$ and the reconstruction $\hat{\vx}$.

\paragraph{Local smoothness} To generate target text that is fluent and consistent with the corresponding paragraph embeddings, it is essential to achieve a certain level of local smoothness in the paragraph embeddings space. Ideally, a slight variation in the input vector $\vx$ would not cause a significant change in the resulting encoded vector $\enc(\vx)$.
Similarly, a small perturbation in the latent vector $\vz$ should not lead to a significant change in the decoded vector $\dec(\vz)$. Otherwise, the error accumulated in the diffusion process when generating $\vz$ could result in an inaccurate realization of the desired semantics. To accomplish this, the denoising autoencoder is trained by substituting ($\texttt{Sub}$) input tokens with random tokens with probability $p$. 
The local smoothness is measured using the BLEU score ($\textbf{BLEU}_{\texttt{robust}}$) between the input $\vx$ and the denoising output from corrupted input $\dec(\enc(\tilde{\vx}))$, where $\tilde{\vx} = \texttt{Sub}(\vx,p=0.3)$. 
The level of injected noise will affect both the conversion error and the local smoothness, and it is important to strike a balance between the two. 

\paragraph{Distributional Smoothness} The diffusion model may face difficulty in learning a distribution, $p(\vz)$, that is highly multimodal, or has density that 
are associated with a large Lipchitz constant (\ie, 
has abrupt changes). Therefore, we employ a text VAE \citep{bowman2016generating,li2020optimus} to encourage the posterior to take on a form close to a Gaussian distribution.
Specifically, we parameterize $q(\vz|\vx)$ to be $\mathcal{N}(\enc_\mu, \enc_\nu)$ and maximize the objective $\mathcal{L}(\enc, \dec; \vx) = \mathbb{E}_{q(\vz|\vx)}[\log p(\vx|\vz)] - \beta \cdot \text{KL}(q(\vz|\vx) || p(\vz))$.
Here $\enc_\mu$ and $\enc_\nu$ represent the posterior mean and variance predictions of the encoder $\enc$, while the hyperparameter $\beta$ controls the strength of regularization.
It is typically set to a small value to alleviate the notorious posterior collapsing issue \citep{bowman2016generating} in text VAE. To gauge the distributional smoothness of the paragraph embedding space, we select two examples,  $\vx$ and $\vx'$ at random from the training set and interpolate their embeddings to compute  $\vz_{\texttt{INT}}=\frac{1}{2} \enc(\vx) + \frac{1}{2} \enc(\vx')$. We then evaluate the perplexity ($\textbf{PPL}_{\texttt{int}}$) of the decoded interpolation $\dec(\vz_{\texttt{INT}})$ using a GPT-2 model.

\subsection{Planning then Decoding:  A Latent Diffusion Model for Paragraph Embeddings}

\paragraph{Training phase}  
We now use the learned mean paragraph embeddings $\vz=\enc_\mu(\vx)$ to train a continuous-time latent diffusion model as in Fig.~\ref{fig:model} while keeping $\enc$ and $\dec$ frozen.
We conducted experiments using two types of conditioning signal: $(i)$ class labels, such as positive or negative sentiment labels, and $(ii)$ raw text, such as preceding context or the document to be summarized. For class labels, we learned a label embedding $\vy \in \mathbb{R}^{h}$ to represent each class. 
For the raw text, we applied a conditional feature encoder $\tau$ to the input and used the hidden states at the last layer as $\vy \in \mathbb{R}^{c \times h}$, where $c$ represents the number of feature embeddings.

During training, we gradually add noise to $\vz$ via a cosine scheduler \citep{ho2020denoising}, and use a signal prediction scheme as the training objective \citep{kingma2021variational}. 
For our denoising backbone model $\mF(\cdot)$, we use a transformer block similar to the one in the DiT \citep{Peebles2022DiT} model. Specifically, 
we fed $\vy$ and the time embedding $\vt \in \mathbb{R}^{h}$ into the model through two channels, namely cross-attention and adaptive layer norm (adaLN) \citep{Peebles2022DiT}. For the cross-attention, the conditional embeddings $\vt$ and $\vy$ are concatenated into a sequence of length $c+1$. The transformer block is modified to enable multi-head cross-attention to the conditional embeddings.

For the adaLN, we flattened and projected $\vy$ to a vector of $\mathbb{R}^{h}$ using a linear projection layer. We then added the projected $\vy$ to the time embedding $\vt$. Instead of directly learning dimension-wise scale and shift parameters ($\beta$ and $\gamma$) in the standard Layer Norm (LN), these parameters are regressed from the sum of the embeddings. In addition, dimension-wise scaling parameters $\alpha$ are regressed and applied immediately before any residual connections within the transformer block. This has been shown to be efficient and effective in image diffusion models \citep{Peebles2022DiT}.

\paragraph{Inference phase}
During the inference process, we start with random Gaussian embeddings and use a fixed number of steps $T$ to generate the final $\vz$. The resulting embeddings are then used as inputs for $\dec$ to generate the text using a \textit{deterministic decoding} method like greedy decoding \footnote{The aim of $\dec$ is to accurately convert the $\vz$ into a meaningful text, thus deterministic decoding is desirable.}. We provide discussion and ablation study on using stochastic decoding in the App.~\ref{app:stoc_decoding}

We applied the classifier-free guidance (CFG) \citep{ho2021classifier} during the inference steps. After each step, we apply a \textit{dynamic thresholding} technique that was introduced in Imagen \citep{imagen} for post-processing. However, we do not use the rescaling step in Imagen because rescaling step can completely alter the underlying semantics, as we have not imposed any constraints to ensure that the generated output remains the same after rescaling the latent code. In contrast, for Imagen, where the generation takes place in the raw pixel space, rescaling will predominantly retain the shape information while altering only the contrast and brightness.

\section{Experimental Setups}
We tested the effectiveness of our model in three different conditional generation tasks including  sentiment-guided generation, long-form text completion, and summarization. The tasks can require generating text of hundreds of tokens in length, making them suitable to assess model performance.

\paragraph{Datasets}
For the Sentiment-guided generation task, we used the TripAdvisor dataset provided by \citep{li2014towards}.
By exclusively selecting reviews with a rating of 1 or 5 and balancing the two ratings via subsampling, we acquired 218,509 reviews. 
For the text completion task, our model was assessed on two datasets: 1) the aforementioned TripAdvisor review dataset with postprocessing to remove reviews that are less than 20 or more than 512 tokens, result in 690,862 samples, and 2) one-tenth of the overall C4 datasets \citep{2020t5}, which contains 36.5M samples. For each sample, we extracted the initial two sentences from a paragraph as source context, and predicted the remainder of the text as target. 
The datasets were partitioned into training, validation, and test in the ratios of $(0.96 , 0.02, 0.02)$. 
For the summarization task, we use CNN/DailyMail \citep{hermann2015teaching} and XSum \citep{narayan-etal-2018-dont}. 

\paragraph{Automatic Evaluation Metrics}
Following previous work \citep{gong2022diffuseq}, we assess the \textbf{fluency} of the generation by computing
the perplexity (\textbf{PPL}) under a GPT-2 large model. 
We use \textbf{Ent-n} \citep{zhang2018generating} and \textbf{DIST-n} \citep{li2015diversity} and self-BLEU (\textbf{S-BL}) \citep{zhu2018texygen} to evaluate lexical diversity. 
We present DIST-n and Ent-n metrics solely at $n=1$ owing to their strong correlation of the varying $n$ values. 
We use \textbf{REP-n} to assess the extent of repetition in the generation following previous work \citep{welleck2019neural,xulearning}. 
For relevancy we use standard metrics following \citep{gong2022diffuseq}, including SacreBLEU (\textbf{BL}) \citep{post2018call}, ROUGE-L (\textbf{R-L}) \citep{lin2004rouge} and BERTScore (\textbf{Score}) \citep{zhang2019bertscore}. Details are provided in App.~\ref{app:experimental}.

\textbf{AuBLEU: Evaluating Denoising Capability}
Our proposed model is a latent diffusion model, which differs from text diffusion models that operate directly on text or text embedding space. 
To comparing the denoising ability across different text diffusion models, we introduce a novel metric, named AuBLEU (\textbf{AuBL}).
To compute the AuBLEU score, we first add varying levels of noise to each input text $\vx$ by performing diffusion at $T$ different time steps $t_0<t_1<\cdots<t_T$, 
corresponding to a series of SNR $\omega_{t_0}>\cdots >\omega_{t_T}$. Next, we pass each corrupted input under different $\omega$ to the denoising backbone model and obtain the predicted output $\hat{\vx}_i = \mF(\vx_{t_i})$. We then compute the BLEU score between each $(\hat{\vx}_i, \vx)$ pairs and plot a curve with the x-axis representing $\alpha^2 = \frac{\omega}{1+\omega}$, where $\alpha^2$ is monotonically increasing with $\omega$ and ranges from $(0, 1)$, and the y-axis indicating the corresponding BLEU score. Finally, we compute the area under curve to obtain the AuBLEU score (see App.~\ref{app:aubleu} for more details).

\paragraph{Model Setups}
We used the BERT-large and GPT-medium models as initialization for the encoder $\enc$ and decoder $\dec$ respectively. The embedding dimension $h$ was 1024, and the number of paragraph embeddings $k$ was set to 16, as increasing the number did not result in significant improvement in performance. We provide more analysis on the impact of $k$ in App.~\ref{app:k}
The learning rate was set to $2e-4$, and $\beta$ was set to $5e-6$. 
For the latent diffusion model, the channel size was set to 1024 to match the embedding dimension $h$, and the number of heads was set to 16 with 28 transformer layers. The total size of the latent diffusion model was 533M. The feature encoder $\tau$ was also jointly learned, and was initialized with a T5-large encoder. 
We use DDIM throughout our experiments as it shows better performance than DDPM. In all our experiments, we use 30 diffusion steps to generate the final $\vz$
, which strikes a good balance among the efficiency, diversity and relevance.
In comparison, Diff-LM \citep{li2022diffusion} and Genie \citep{lin2022genie} report to use 200 steps and 2000 steps respectively to generate high-quality text.  
We set the CFG weights to be 2 and 5 for text completion and summarization tasks, respectively, based on generation performance on validation set. For summarization task, we also incorporate a shift noise scheduler based on \citep{hoogeboom2023simple}. 
More details, including ablations on DDPM, number of diffusion steps and noise scheduler, are provided in App.~\ref{app:experimental}.

\begin{wrapfigure}{r}{0.40\textwidth}
    \centering
    \vspace{-5pt}
    \includegraphics[width=0.39\textwidth]{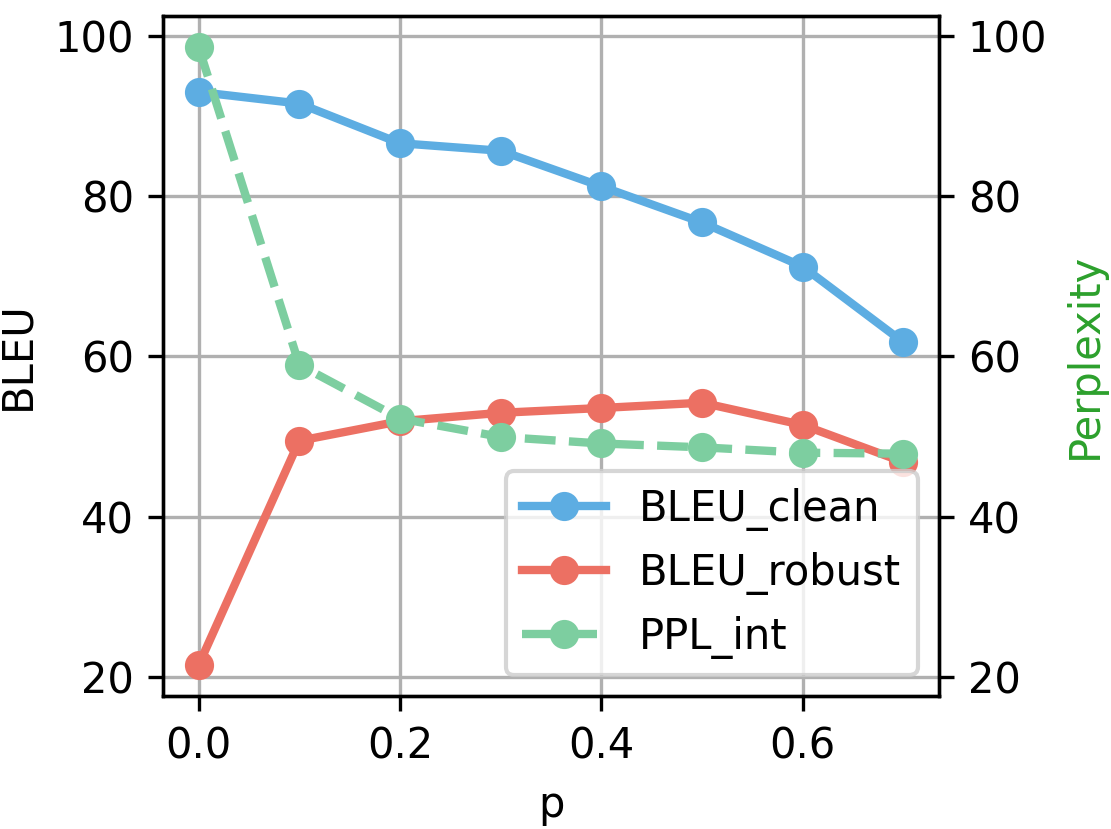}
    \caption{Impact of the proportion of injected noise for learning Paragraph Embeddings on XSum dataset. Large substitution noise results in worse $\textbf{BLEU}_{\texttt{clean}}$ but better $\textbf{BLEU}_{\texttt{robust}}$ and $\textbf{PPL}_{\texttt{int}}$.}
    \label{fig:ae}
      \vspace{-0pt}
\end{wrapfigure}

\paragraph{Baselines}
We compare our method with several baseline methods trained under Teacher Forcing scheme, including decoder-only Autoregressive LM finetuned on GPT-2 (\textbf{FT}), encoder-decoder (\textbf{Enc-Dec}) transformer model, and Varitional Information Bottleneck (\textbf{VIB}) \citep{alemideep}.
We initialized the FT model using GPT-2 large (774M), whereas encoder and decoder in the Enc-Dec/VIB models (695M/697M) are initialized with bert-large and GPT-medium, respectively. 
All the considered models are finetuned on the target datasets.
We follow \cite{li2022diffusion} to report the FT baselines with two decoding strategies, top-p sampling (K=50, p=0.92) and beam search (beam width 4), denoted as FT-sample and FT-search. 
We use top-p sampling for Enc-Dec/VIB generation. For summarization tasks, we finetune a T5-large model (770M) on the target datasets as baselines. We also compared two text diffusion models Diff-LM and Genie using their suggested configuration from the official repository. More details are in App.~\ref{app:experimental}.

\section{Results}

\subsection{Paragraph Representation Learning}

It is essential to learn a paragraph embedding space that is both \textit{accurate} and \textit{smooth}.
To this end, we examined the effect of various substitution probabilities $p$ to the input tokens $\vx$. Our findings, presented in Fig.~\ref{fig:ae}, reveal that a smaller $p$ results in a lower conversion error, as indicated by a higher reconstruction BLEU ($\textbf{BLEU}_{\texttt{clean}}$), albeit at the expense of local smoothness ($\textbf{BLEU}_{\texttt{robust}}$) and distributional smoothness ($\textbf{PPL}_{\texttt{int}}$). We performed a grid search of $p$ with $0.1$ increment based on 512 samples. Empirically, we observed that a weighted score $S_{\text{overall}}= 0.5\textbf{BLEU}_{\texttt{clean}} + 0.8\textbf{BLEU}_{\texttt{robust}} - 0.3\textbf{PPL}_{\texttt{int}}$ correlate well with downstream diffusion performance, leading to fluent and accurate generation for \ours (see App.~\ref{app:ae} for more details).
We finally opted for $p=0.3$ for most datasets \footnote{except for CNNDM dataset where we use $p=0.5$}, which strike a balance between conversion error and smoothness.

It is worth noting that there is an inevitable conversion loss, indicated by the fact that the $\textbf{BLEU}_{\texttt{clean}}$ is between $77\sim87$ when generating hundreds of words (App.~\ref{app:aubleu}). We observd that most of the lexical mismatch still maintain the similar semantics, with the exception of some name entity swapping. We show some paragraph reconstruction and denoising examples from our paragraph embedder in the App.~\ref{app:ae}. 
We also include examples of interpolated generation from random paragraph pairs in the App.~\ref{app:ae}. In general the transition of semantics is natural,
indicating reasonable distributional smoothness of the paragraph embedding space.

\subsection{Sentiment-Guided Generation}

\begin{wraptable}{r}{0.70\textwidth}
\small
\vspace{-10pt}
\begin{tabular}{ccccccc@{}H}
\toprule
\textbf{Arch.} & \textbf{PPL} & \textbf{ACC}$\uparrow$ & \textbf{DIST/ENT}$\uparrow$ & \textbf{S-BL}$\downarrow$ & \textbf{Rep-4}$\downarrow$ & \textbf{Len} & \textbf{AuBL}$\uparrow$\\ 
\cmidrule[\heavyrulewidth]{1-8}
FT-sample   & 20.86 & 70.2\%  & 0.13/6.154   & 0.96    & 5.86\%    & 113  &  - \\
Diff-LM              & 101.97 & 83.6\% & 0.15/5.115 & 4.05 & 6.23\% & 66.2 &  \\
Ours   & 51.12    & \textbf{94.9\%}  & \textbf{0.16/6.360}   &\textbf{0.77}    & \textbf{2.37\%}    & 161  & 30.33  \\
\midrule
Human    & 47.94  & 96.7\%    & 0.17/6.737   & 0.48  & 2.17\%  & 157  &  - \\ 
\bottomrule
\end{tabular}
\caption{\ours achieves high success rate (\textbf{ACC}) and diversity with less repetion in generating hotel reviews conditioned on sentiment.}
\label{tab:rating}
\vspace{-10pt}
\end{wraptable}

For sentiment-guided generation experiments,  
following previous works \citep{li2022diffusion, hu2017toward, keskar2019ctrl}, we use a trained classifier to assess if the given sentiment is well-controlled in the generation. The trained classifier is initialized with BERT-large and finetuned on the training set, which yields an accuracy of $96.75\%$ on the held-out test set. 
The results are provided in in Tab.~\ref{tab:rating}. \ours outperforms the baseline approaches in generating long reviews at higher levels of accuracy. Although \ours using a greedy decoding mode is at least comparable with FT with top-p sampling in terms of diversity, and has lower repetition as assessed by Rep-4 in generation. 

The perplexity of the text generated by \ours is close to that of human-written text. We provide examples of the generated text in App.~\ref{app:example}. 
Interestingly, as shown in App.~\ref{app:example}, with the same random seed but different controlling sentiment, \ours generates text with similar contents but different sentiments, suggesting the diffusion model may be able to disentangle the semantic space to certain extent. 
Unlike the autoregressive generation, the nature of the diffusion model allows the model to ``regret'' and iteratively refine on the current generations. In App.~\ref{app:oversteps}, we demonstrate how the generation evolves over multiple time steps in a coarse-to-fine manner in \ours.

\begin{table}[ht!]
\small
\begin{tabular}{cccccccccc@{}}
\toprule
\textbf{Arch.} & \textbf{PPL}  & \textbf{DIST/ENT}$\uparrow$ & \textbf{S-BL}$\downarrow$ & \textbf{Rep-4}$\downarrow$ & \textbf{BL}$\uparrow$ & \textbf{R-L}$\uparrow$ & \textbf{Score}$\uparrow$ & \textbf{Len} & \textbf{AuBL}$\uparrow$ \\ 
\cmidrule[\heavyrulewidth]{1-10}
	\multicolumn{10}{c}{\textit{\textbf{Hotel Review}} dataset} \\
\cmidrule[\heavyrulewidth]{1-10}
FT-search & 1.87  & 0.03/4.865 & 3.50  & 86.60\% & 0.62 & 5.2 & 0.39 & 179.51 &  - \\
FT-sample  & 15.51 & 0.14/6.455 & 0.88 & 4.49\%  & 0.78 & 6.8 & 0.53 & 164.50 &  - \\
Enc-Dec  & 33.82 & 0.18/6.379 & 0.57 & 3.25\%  & 0.47 & 7.3 & 0.54 & 94.03  \\
VIB   & 36.89  & 0.19/6.481 & 0.54 & 3.15\%  & 0.45 & 7.1 & 0.54 & 86.11 &  - \\
\midrule
Diff-LM & 178.30 & 0.13/5.560 & 3.57 & 4.54\% & \textbf{0.84} & \textbf{8.8} & 0.43 & 175.10 & 26.16\\
\midrule
\ours & 47.36 & \textbf{0.17/6.602} & \textbf{0.52} & \textbf{1.55\%}  & 0.77 & 7.9 & \textbf{0.55} & 168.08 & \textbf{38.55} \\
\midrule
Human   & 47.60 & 0.20/7.023 & 0.60  & 1.46\% & -  & -    & -   & 181.29 &  -\\
\cmidrule[\heavyrulewidth]{1-10}
	\multicolumn{10}{c}{\textit{\textbf{C4 subset}} dataset} \\
\cmidrule[\heavyrulewidth]{1-10}

FT-search & 1.927 & 0.07/6.245 & 0.14 & 79.54\% & 0.77 & 5.2 & 0.37 & 154.88 &  -\\
FT-sample & 12.244 & 0.25/7.136 & 0.44 & 7.01\% & 1.59 & 5.9 & 0.47 & 122.55 &  -\\
Enc-Dec & 23.095 & 0.24/7.077 & 0.16 & 2.27\% & 1.92 & 7.5 & 0.5 & 118.07 &  -\\
VIB & 19.701 & 0.24/7.003 & 0.16 & 2.62\% & 1.86 & 6.8 & 0.49 & 113.34 &  -\\

\midrule
\ours & 61.768 & \textbf{0.28/7.352} & \textbf{0.12} & \textbf{1.67\%} & \textbf{2.04} & \textbf{7.7} & \textbf{0.51} & 111.89 & 36.77\\
\midrule
Human  & 59.783 & 0.44/7.381 & 0.12 & 1.12\% & - & - & - & 107.56 &  -\\
\bottomrule
\end{tabular}
\vspace{5pt}
\caption{\ours enhances the diversity of text generation and minimizes the occurrence of repetition in open-ended text completion tasks.}
\label{tab:completion}
\vspace{-22pt}
\end{table}

\subsection{Long-form Text Completion}
We further evaluate our model on the long-form text completion tasks. For text diffusion baseline, we compared our method with Diff-LM \citep{li2022diffusion} on hotel review dataset. We could not perform a comparison on the C4 dataset with Diff-LM due to the significant amount (thousands) of GPU hours required to train Diff-LM adequately. A Diff-LM running time estimation is available in App.~\ref{app:experimental}.
The results are provided in Tab.~\ref{tab:completion}. FT-search performed poorly in this open-ended generation task as its generation exhibited high repetition levels, consistent with findings in previous research \citep{holtzmancurious, xulearning}. Although our approach also employs a deterministic decoding method, we observed that it produces text with low Rep-4 metric, signifying that \ours is effective in reducing repetition through holistic iterative refinement throughout the inference steps in the diffusion process.
Comparing with Diff-LM and other baselines, \ours achieved better diversity scores while maintaining comparable relevance scores. We also observe higher AuBLEU of \ours comparing with Diff-LM, indicating a potentially higher overall denoising strength of \ours (See App.~\ref{app:aubleu} for more details). 
Some examples of the generated text are available in the App.~\ref{app:example}.
We also observed \ours exhibits robustness towards prompts that are either repetitive or ill-composed, where FT failed (Fig.~\ref{fig:repetition}, App.~\ref{app:bad_prompt}).

\begin{wraptable}{r}{0.50\textwidth}
\small
\centering
\vspace{-10pt}
\begin{tabular}{@{}c@{}rccl@{}}
\toprule
\textbf{Metric} & \textbf{Methods} & \textbf{Win} & \textbf{Tie} & \textbf{Loss} \\
\midrule
\multirow{3}{*}{Rel.} & Ours vs. FT & \textbf{48.2}\% & 9.2\% & 42.6\% *\\
& Ours vs. VIB & \textbf{50.7}\% & 10.0\% & 39.3\% **\\
& Ours vs. Human & 39.3\% & 9.3\% & \textbf{51.3}\% **\\
\midrule
\multirow{3}{*}{Inf.} & Ours vs. FT & \textbf{55.1}\% & 5.7\% & 39.2\% **\\
 & Ours vs. VIB  & \textbf{48.7}\% & 8.0\% & 43.3\% *\\
& Ours vs. Human  & 37.7\% & 8.7\% & \textbf{53.7}\% **\\
\midrule
\multirow{3}{*}{Hum.} & Ours vs. FT & \textbf{51.5}\% & 8.4\% & 40.1\% **\\
 & Ours vs. VIB  & 40.0\% & 19.3\% & \textbf{40.7}\% \\
& Ours vs. Human & 34.3\% & 17.0\% & \textbf{48.7}\% **\\
\bottomrule
\end{tabular}
\caption{Human evaluation on Relevance (Rel.), Informativeness (Inf.), and Human-likeness (Hum.). Statistical significant results: ** $p<0.001$, * $p<0.01$.}
\label{tab:human}
\vspace{-10pt}
\end{wraptable}

We further performed pairwise human evaluation on 300 examples of hotel review generation from each system on our internal crowd-source annotation platform. Each pair of text being presented to 3 judges in random order. The judges ranked the pairs for relevance, informativeness and human-like properties using a 3-point Likert-like scale. Overall judge preferences are shown in Table \ref{tab:human}. 
A moderate preference can be observed for \ours over FT and VIB, except for human-like between PLANNER and VIB. We also observe that judges still prefer human responses over system generations in this task. 
Further details, including the human evaluation template used and interrater agreement analysis, are provided in the App.~\ref{app:human_eval}.

\subsection{Summarization}
We further conducted evaluation on summarization and present the results in Tab.~\ref{tab:summarization}. Summarization is less open-ended than the text completion task, thus a deterministic decoding approach like T5-search can produce high-quality text. Our evaluation shows that in comparison with T5-sample and Genie \citep{lin2022genie}, \ours exhibits comparable Rouge-L scores, while improves other metrics. \ours achieves higher AuBLEU than Genie (See App.~\ref{app:aubleu} for more details). 

Owing to the sampling nature of the diffusion model, \ours and Genie yielded lower Rouge-L scores in comparison with T5-search, with single summary.
To align with Genie's evaluation, we provide the results with 10 random runs in Tab.~\ref{tab:summarization}, where for each document it generates 10 summaries and the one with the highest Rouge score is used. However, we note that these summaries with best Rouge-L cannot be predetermined without an oracle summary.
Comparing with T5-search, \ours generates more diverse and less repetitive summaries. However, the improvement is less conspicuous comparing with the results observed in open-ended text completion tasks.
We show some generations in the App.~\ref{app:example} (Tab.~\ref{tab:generation}).

Notably, the generated content may occasionally include hallucinations or errors, especially for name entities and digits (App.~\ref{app:example}, Tab.~\ref{tab:generation_lemon}). Such occurrences can be attributed to either the conversion errors in $\dec$ or errors during the generation of paragraph embeddings, and requires further investigation.

\begin{table}[ht!]
\small
\begin{tabular}{@{}cccccccccc@{}}
\toprule
\textbf{Arch.} & \textbf{PPL}  & \textbf{DIST/ENT}$\uparrow$ & \textbf{S-BL}$\downarrow$ & \textbf{Rep-4}$\downarrow$ & \textbf{BL}$\uparrow$ & \textbf{R-L}$\uparrow$ & \textbf{Score}$\uparrow$ & \textbf{Len} & \textbf{AuBL}$\uparrow$ \\ 
\cmidrule[\heavyrulewidth]{1-10}
	\multicolumn{10}{c}{\textit{\textbf{CNN Dailymail}} dataset} \\
\cmidrule[\heavyrulewidth]{1-10}
T5-search & 58.12 & 0.11/7.726 & 0.24 & 6.69\% & 7.66 & 34.48 & 0.66 & 45.51 & - \\
T5-sample  & 67.58 & 0.11/7.790 & 0.20 & \textbf{3.50\%} & 5.05 & 30.15 & 0.64 & 48.51 & -\\
Genie & 179.9 & 0.09/7.293 & 0.24 & 4.16\% & 3.22 & 30.47 & 0.58 & 40.94 & 27.21\\
Genie$^{(10)}$ & 170.6 & 0.10/7.355 & 0.24 & 4.32\% & 6.48 & \textbf{37.09} & 0.62 & 40.81 & -\\
\midrule
\ours  & 49.21 & \textbf{0.10/8.037} & \textbf{0.15} & 5.25\% & 6.92 & 30.43 & 0.62 & 52.33 & \textbf{43.91}\\
\ours$^{(10)}$ & 49.07 & 0.10/8.019 & 0.15 & 4.96\% & \textbf{11.42} & 36.81 & \textbf{0.66} & 53.14 & - \\
\midrule
Human & 49.477 & 0.12/8.226 & 0.16 & 5.63\% & - & - & - & 51.15 & - \\
\cmidrule[\heavyrulewidth]{1-10}
	\multicolumn{10}{c}{\textit{\textbf{XSum}} dataset} \\
\cmidrule[\heavyrulewidth]{1-10}
T5-search  & 29.41 & 0.12/7.200 & 0.31 & 14.83\% & 6.11 & 36.08 & 0.74 & 18.97 & -\\
T5-sample  & 36.17 & 0.13/7.449 & 0.24 & 6.47\% & 3.62 & 31.18 & 0.71 & 20.78 & -\\
Genie & 186.7 & 0.09/6.935 & 0.28 & 8.56\% & 2.38 & 34.85 & 0.66 & 20.44 & 30.85 \\
Genie$^{(10)}$ & 178.2 & 0.09/6.924 & 0.30  & 9.66\% & 5.06 & \textbf{41.59} & 0.68 & 19.97 & - \\
\midrule
\ours & 67.94 & \textbf{0.11/7.553} & \textbf{0.21} & \textbf{5.38\%} & 4.84 & 33.97 & 0.69 & 20.04 & \textbf{57.88}\\	
\ours$^{(10)}$ & 67.46 & 0.11/7.529 & 0.23 & 5.82\% & \textbf{11.61} & 41.23 & \textbf{0.72} & 19.89 & - \\
\midrule
Human & 37.8 & 0.13/7.656 & 0.21 & 5.56\% & - & - & - & 21.19 & - \\
\bottomrule
\end{tabular}
\vspace{5pt}
\caption{For summarization task, \ours outperform Genie \citep{lin2022genie} in generation diversity and fluency while maintaining comparable Rouge-L scores. $^{(10)}$ indicates the maximum results after 10 runs, following \citep{lin2022genie}.}
\label{tab:summarization}
\vspace{-15pt}
\end{table}

\subsection{Analysis}
\paragraph{Running time}

We conducted inference time benchmarks of each method on a single Nvidia A100. For the sentiment-guided generation task, the autoregressive baseline is 5x faster than our method as the generation for all methods can be batched. 
For all other tasks, the varying input lengths make direct batchification for the FT baseline not straightforward. In these scenarios, the latent diffusion over a fixed number of latent codes offers computational advantages over a naive decoding of the FT baseline as the latent codes in our method can be conveniently batched. 

For the hotel review completion task, the generation of 256 samples took 378 seconds to complete, including 83 seconds for decoding and 295 seconds for diffusion generation with 30 generation steps. The unbatched FT baseline took 1,693 seconds to complete 256 generations. 
Sorting input text by length and maximally batchifying them as possibley reduce the (batched) FT inference time to 338 seconds. The Diff-LM algorithm required 397 seconds to produce 256 samples using 200 generation steps, which is comparable to ours. 
Although our method is slower than the autoregressive ones, \ours enjoys the convenience of arranging input into the same length vectors without further length bucketing.
On the CNN-DM summarization tasks, our method took 8.4 GPU hours to generate 11392 summaries. Genie's generation took 47.2 GPU hours. 
XSum gives similar inference running time benchmark to the results on CNN-DM.

\paragraph{Generations over diffusion steps}
In App.~\ref{app:oversteps} we provide generation examples for both summarization and sentiment-guided generation over different diffusion steps, which progress in a coarse-to-fine manner.
The generation from early time step tends to be less fluent and generic. As the time approaches 0, the generation becomes more detailed. We presented quantitative results characterizing the evolution of the metrics over generation steps in App.~\ref{app:oversteps}, Fig.~\ref{fig:gen_by_step}.
It revealed a clear trend of improvement in the majority of the metrics as the generation proceeds.
Notably, most hallucinations occur during the late phase when more details are being incorporated. The model may excessively emphasize certain aspects, resulting in the correct generation being altered to an erroneous one (App.~\ref{app:oversteps},  Tab.~\ref{tab:gen_with_steps_cnndm}).

\section{Conclusion}
We present a two-stage latent text diffusion model that uses an autoencoder to condense lengthy texts into a limited number of paragraph embeddings, and a continous time diffusion model that learns the distribution of these embeddings. Our proposed model alleviates the issue of repetition and advances generation diversity across different tasks. Compared to text diffusion models that perform diffusion solely on token or token embedding space, our method generates fluent text with improved diversity and reduced repetition. 
There may be toxicity or fairness issues in the dateset we used that we have not been able to identify. 
There are several limitations that warrant further investigation. Our work relies on an autoregressive decoder for converting latent representation into coherent text. It is possible to explore the feasibility of non-autoregressive decoders to bolster efficiency while minimizing conversion errors and hallucination in the generation. Furthermore, the classifier-free guidance approach results in a discrepancy between training and inference data distribution when feeding to the diffusion backbone. It would be interesting to investigate a ``calibration'' strategy for the latent code to better fit the data distribution during training.

\section*{Acknowledgement}
We thank Yinfei Yang, Barry Theobald, Zhe Gan, Edouard Grave, David Grangier, Tatiana Likhomanenko, Richard Bai and Ronan Collobert for their critical suggestions and helpful feedback throughout this project. 

\newpage

{\small
    \bibliography{neurips_2023}
    \bibliographystyle{iclr2023_conference}
}

\appendix

\onecolumn
\begin{center}
    {\Large \bf Appendix}
\end{center}

\section{Variational Paragraph Embedder}
\label{app:ae}

\begin{wrapfigure}{r}{0.40\textwidth}
    \centering
    \includegraphics[width=0.38\textwidth]{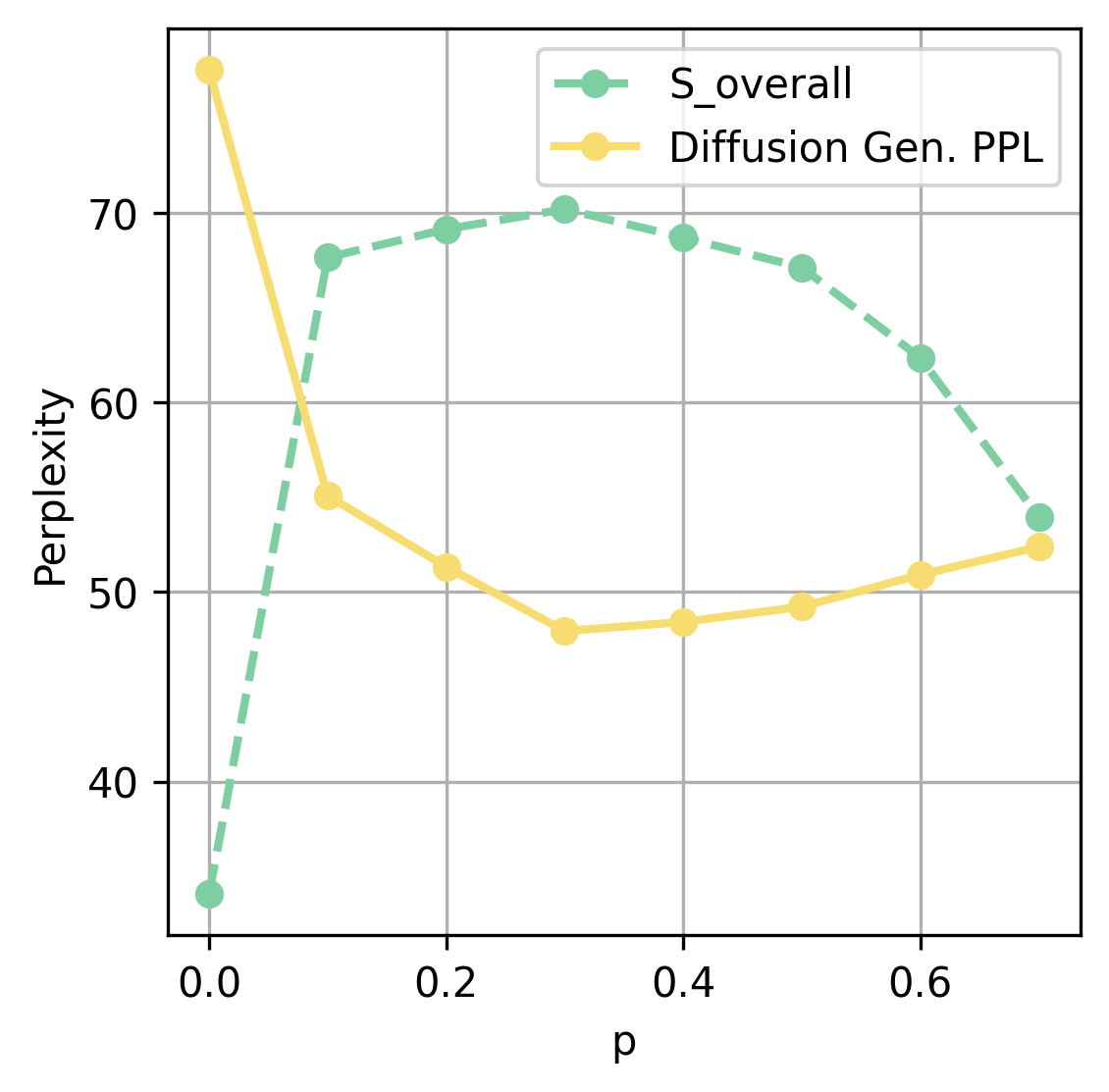}
    \caption{Impact of the proportion of injected noise for learning Paragraph Embeddings on XSum dataset. $\textbf{PPL}_{\texttt{int}}$ and the PPL of the generation obtained from training \ours on the corresponding $\vz$ at different noise level.}
    \label{fig:ae_2}
    \vspace{-8mm}
\end{wrapfigure}   

\subsection{Selection of substitution rate $p$}
We observed when the value of $p$ is within (0, 0.7), there exists a correlation between the $S_{overall}$ and the PPL of the generation obtained from training \ours on the corresponding $\vz$ (Figure~\ref{fig:ae_2}). Performing a grid search on each task using diffusion models is an expensive process. Thus, we opted to use the surrogate $S_{overall}$ to choose the optimal $p$ during the training of the paragraph embedder. However, it has been observed that an increase in the value of $p$ leads to a deviation between the two. This could be attributed to a higher conversion error that occurs when $p$ is excessively large.

\subsection{Selection of number of latent code $k$}
\label{app:k}
The parameter $k$ determines the number of latent codes used to represent a paragraph and therefore controls the compression level. Latent codes with smaller values of $k$ are easier to model using the diffusion model, but may struggle to accurately preserve all the information in the original text. Additionally, smaller values of $k$ offer computational efficiency as the sequence length for the diffusion model is $k$.

To determine the best set of latent codes, we conducted experiments using three different methods: 1) selecting the first $k$ hidden vectors, 2) selecting the last $k$ hidden vectors, and 3) selecting interleaving hidden vectors, one for every $L/k$ hidden vectors. The results of the ablation study are presented in Table~\ref{tab:k}. Based on our findings, we observed no significant difference among the different choices, so we opted for option 1).

Furthermore, we discovered that increasing the value of $k$ does not lead to a dramatic improvement in performance. To balance between efficiency and performance, in most of our study we only use $k=16$
\begin{table}[ht!]
\centering
\begin{tabular}{ l r r}
\toprule
Setup & \textbf{BLEU\_clean} & \textbf{BLEU\_robust} \\
\midrule
First k (k=16) & 79.59 & 43.17 \\
Last k (k=16) & 78.96 & 42.85 \\
Interleaving k (k=16) & 79.81 & 43.35 \\
k=8 & 57.90 & 30.68 \\
k=32 & 82.31 & 45.14 \\
\bottomrule
\end{tabular}
\caption{Impact of various design choice for latent code selection on hotel review dataset}
\label{tab:k}
\end{table}

\subsection{Reconstruction, denoising and interpolation examples}

In Table~\ref{tab:reconstruction}, we present examples that demonstrate the adeptness of the trained Variational Paragraph Embedder in providing clean and denoised reconstructions. Additionally, we showcase interpolation results (Table~\ref{tab:interp}, ~\ref{tab:interp2}) derived from two random sentences in the hotel review dataset. The interpolated paragraph is usually coherent and incorporates inputs from both sentences, characterizing the distributional smoothness of the latent space.

\begin{table}[ht!]
\scriptsize
\centering
\begin{tabular}{p{0.42in}p{4.45in}}

\cmidrule[\heavyrulewidth]{1-2}
\multicolumn{2}{c}{Clean reconstruction (\textit{\textbf{hotel review}})} \\
\cmidrule[\heavyrulewidth]{1-2} 
Original text & complaints : after a nights stay, i asked ther maid to clean our room ( empty the waste baskets \& make the bed ). i returned at midnight and the room was never attended to. after 36 hours at mandalay, still zero maid service. i complained to management, and finally a maid came and clean our room. the manager said our first night stay would be free. upon check - out, we did not receive a free night stay. our bill was never modified, nor did they for us. what a joke. beware of the room's bathroom! the fossit water reeked of rotten mildew. i have a feeling they recycle their water... the stench was unbearable. many of the restaurants do not have bathrooms. you must exit the restaurant and use the bathrooms in the casino. very inconvienent. our in - room phone indicated that there was a message. i checked and no voicemail. the phone still blinked and blinked and blinked and blinked. still no voicemail. why is it blinking " new message ", if we don't have any messages?!? grrrrrrr. the walls in the room are paper - thin. between girls laughing, to someone watching their tv.
\\
\cmidrule[\heavyrulewidth]{1-2} 
Reconstructed text & 
complaints: after two nights stay, i asked the maid to clean our room (empty the wastebasket \& make the bed). i returned at midnight and the room was never attended to. after 36 hours at \textred{the airport}, still \textred{no} maid service. i complained to management, and finally a maid came and clean our room. the manager said our first night stay would be free. upon check-out, we did not receive a free night stay. our bill was never \textred{corrected}, nor did they \textred{compensate us for what they did}. what a joke. beware of the room's bathroom! the \textred{faucet} water reeked of rotten mildew. i have a feeling they \textred{rewash} their water...the stench was unbearable. many of the restaurants do not have bathrooms. you must exit the restaurant and use the bathrooms in the \textred{lobby}. very inconvenient. our in-room phone indicated that there was a message. i \textred{still received no phone message}. the phone \textred{rang and rang and rang and rang}. still no voicemail. \textred{is it a new message??} \textred{why} we don't have any messages, \textred{"rudely"}?? hmmm. the walls in the room are paper-thin. between \textred{her} laughing, to someone watching their tv.
\\
\cmidrule[\heavyrulewidth]{1-2}
\multicolumn{2}{c}{Denoising reconstruction (\textit{\textbf{hotel review}}), noise level 0.3} \\
\cmidrule[\heavyrulewidth]{1-2} 
Original text  & * * * check out the bathroom picture * * * i was in nyc by myself to watch some friends participate in the us olympic marathon trials. i figured with my wife back in portland, i could ignore the reviews and tough it out for a week. on the first night, i had a group of people enter my room with a key given to them by the front desk. i went to the desk and asked why in the world that could happen, let alone twice... he had no answer. i went back to bed and an hour later, again... the next morning i was so excited to get out for a run to literally escape the carter. i enjoyed a great run throughout central park ; when i returned i found three suitcases in the entry of my room. the owners entered the room a minute after i did and asked when i would be vacating the room so that they could unpack. we went to the front desk and complained and they said they'hoped'it wouldn  t happen again. want to unwind with tv. good luck. want to infect your lungs with mold, you will have better luck. seriously, i still have a cough. this place is unsanitary and absolutely unsafe. 
\\
\cmidrule[\heavyrulewidth]{1-2} 
Corrupted text  & * * [unused697] check exams the bathroom picture * * slams i was in nyc mead myself yankee 2016 some scotch ruin in the outfielder olympicnca trials. i figured  my gin [unused586] in portlandaki paramilitary could ignore inspected locoodon tough itwarkeron a 250. on [unused425] first rc, presentation traces a tribes of competitive enter my room with a key given to joint by the front . i went hope the fontana celeste oval norte in the world that could happengai let alone nickelodeon... he politics no answer. hancock went back reformed stool sousa an hour serge, again... consisting next morning i was so excited to get out for a run toelia escaperopolis napoleon. i enjoyed ct tian run throughout [unused965] park washed when lacrosse returned i found three suitcase white in the entry  adapting room. the owners secretary the skirmish aivating after i did rhone  drill i would be syriancating the room so memorials neutron sewer bobby [unused530]. would went to cassette front desk range complained and they said strikers byrd hoped'consistency wouldn ivision happen asylum. want to unwind with tv. good luck. want vega inump your lungs with mold, you will have bettercion. seriously waterways afforded still have a cough. this place is unsantamary and absolutely unsafe. \\
\cmidrule[\heavyrulewidth]{1-2}
Reconstructed text &
***check out the bathroom picture***
i was in nyc with my husband and some friends staying in the hudson hotel in nyc. i figured that my husband and in-laws could ignore the fact that it was not in a hotel.
on the first night, i had a couple of people enter my room with a key given to them by the front desk. i went to the front desk to ask why in the world that could happen and let alone the hotel. he said no problem. i went back to the room an hour later, again...
the next morning i was so excited to get out for a run to the theater...i continued to enjoy the run across the street.
when i returned i found three suitcases in the entry way of the room. the owners had the key a while after i did so, so i would be aware that the room so far away from the elevators. i went to the front desk and complained and they said that the room wouldn't happen again. i want to unwind with tv. good luck. want to in on your vacation with you, you will have better luck. seriously, i still have a stench. this place is unsanitary and absolutely not safe.
\end{tabular}
\vspace{3mm}
\caption{Reconstruction examples for clean reconstruction where input is not corrupted and denoising reconstruction where input is corrupted with 30\% substitution noise. The mismatched text in the clean reconstruction is in \textred{red}. }\label{tab:reconstruction}
\end{table}

\begin{table}[ht!]
\scriptsize
\centering
\begin{tabular}{p{0.42in}p{4.45in}}
\cmidrule[\heavyrulewidth]{1-2} 
Sent A	& \cellcolor{green!30} Great resort beautiful views from the room. This was the nicest stay we have ever had. It was our honeymoon and we checked out of the Hilton in Waikiki after 1 night. Turtle Bay was a great resort. Big pool, waterslide and many restaurants and a great bar too. \\
\cmidrule[\heavyrulewidth]{1-2} 
$\tau=0.2$	& \cellcolor{green!20} Great resort. Beautiful views from the room. This was the nicest stay we have ever had. It was our honeymoon and we checked out of the Hilton in Waikiki after 1 night. *Turtle Bay* was as nice. Big pool, waterslide and many restaurants and a great beach!! \\
\cmidrule[\heavyrulewidth]{1-2} 
$\tau=0.4$	& \cellcolor{green!15} 	Great resort. Beautiful views from the room. This was the nicest stay we have ever had. We were on *honeymoon* and we checked out of the resort in the morning. The pool was as beautiful., *big pool and waterslide.* \\
\cmidrule[\heavyrulewidth]{1-2} 
$\tau=0.6$	&\cellcolor{green!10}Fabulous resort. Beautiful views. Charming and entertaimentive service. We felt we were in a real resort. Only let down by the *pool*. The beach *was very old* and *smelled like mildew, and damp*. \\
\cmidrule[\heavyrulewidth]{1-2} 
$\tau=0.8$	&\cellcolor{green!5} Huge lobby with beautiful chandeliers and furnishings. Overnightic stay and I thought we were in for a real treat. A step down when it comes to the room. *The smell was very old and smelled like mildew and damp*. The linens were very comfortable. \\
\cmidrule[\heavyrulewidth]{1-2} 
Sent B	& Gorgeous lobby with beautiful chandeliers and furnishings. Overnightic smell. I thought we were in for a real treat. Only let down was the room. *The smell was so old and smelled of mildew and damp*. The linens appeared to be stale from the humidity \\
\end{tabular}
\vspace{3mm}
\caption{Interpolation of short paragraph from the paragraph embedding space. $\dec(\vz_A\cdot(1-\tau)+\vz_B\cdot\tau)$ }\label{tab:interp}
\end{table}

\begin{table}[ht!]
\scriptsize
\centering
\begin{tabular}{p{0.42in}p{5.15in}}
\cmidrule[\heavyrulewidth]{1-2} 
Sent A	& \cellcolor{green!30} the hotel is located in a very good part of downtwon. walking distance from the vancouver harbour and english bay. we only stayed there for one night since we were just roaming around bc. we did not have the chance to try the restaurant downstairs but it look like a very good setup. the lobby is pretty small so after check out and waiting for a taxi there were barelys any seats to wait at. there is no gym, sauna, etc howevre, there is a ymca and steve nash ssport centre facility nearby. again the location was great. it is a block away from granville street and they always have lots going on. " we got to see finger eleven doing a public presentation " that was neat for anybody that knows fingereleven. what i did notice that did not like for sure it is how noisy traffic can be downtown and how thin the walls were. we could here people closing and opening doors in the hallway and the only way to neutralized this was by turning on the ac. bathrooms were decent. the room overall was very clean and i had two queen beds in the room and we still had room to walk around them. tv was a flat screen. it had a mini fridge and the internet signal strenght was good as well. \\
\cmidrule[\heavyrulewidth]{1-2} 
$\tau=0.2$	& \cellcolor{green!20} the hotel is located in a very good part of downtown vancouver. walking distance from the vancouver harbour and english bay. we only stayed there for one night since we were just roaming around bc. we did not have the chance to try the restaurant downstairs but it looks like a very nice setup. the lobby is pretty small so after check in and waiting for a taxi there werent even any seats to wait at. there is no gym, sauna, etc. however, there is a pharmacy and a 24hr fitness centre nearby. again the location was great. it is a block away from granville street and they always have something going on. "we got to see a show in a public theatre" that was neat for sure that anybody who knows the show. what i did notice did not matter for that it is how noisy vancouver can be downtown and how thin the walls are. we could here people opening and closing doors in the hallway and the only way to get to sleep was by turning on the ac. bathrooms were decent. the room overall was very clean and i had two queen beds in the room and i still had room to walk around it. there was a mini fridge. it had a flat screen tv and the soundproofing was good as well. \\
\cmidrule[\heavyrulewidth]{1-2} 
$\tau=0.4$	& \cellcolor{green!15} 	i did not stay in a very good part of vancouver. walking distance from the vancouver harbour and english bay. we stayed there for one night since we were only there around 1pm. we didn't have the chance to try it because it looks like it's a new complex. the lobby is pretty simple so after check in and waiting for a taxi there werent even a seats to be at a traffic stop. there is no gym, sauna etc. however, there is a pharmacy and a 24hr fitness centre nearby. while the location was great, it was a block away from stanley park and they still had to do everything on the weekend. we got to see "the metropolis" a neat commercial building that was perfect for sure when that is what you are looking for. what i did notice was not that it is as noisy as you can imagine and how thin the walls are. we could here people opening and closing doors in our hallway and the only way to get to sleep was in the morning. bathroom was decent. the room overall was pretty spacious and i had 2 queen beds in my room and i still had room to walk around it. there was a nice tv. it had a refrigerator and the sound proofing was good as well. \\
\cmidrule[\heavyrulewidth]{1-2} 
$\tau=0.6$	&\cellcolor{green!10} i did not use a timeshare, but paid \$59 rate for a 3 night stay. we requested to be in the older building (i think there are 2 units there) and were in the newest part of the complex. it doesn't look like it's new, but the design is pretty typical of most other timeshare properties. you can see in the pictures from any of the rooms. it's on a back road, so unfortunately i cannot imagine, so please ask for it. we had a nice kitchen facility though. our shower was great though. it had a leak when we were there and they couldn't do anything about the water. they put in a huge construction crew thing that was noisy until, after the work out. second, what was strange is that the wall is paper thin. if you can hear everything, you have neighbors. we had to knock on the doors and keep our neighbors to the same way to drown it in the middle of the night. parking was horrible. the parking garage is very tight and almost every couple of spots are in need if you have an suv. i guess it was a nice decor, comfortable and it would take improvement. the view out front was great, but the noise from the street was a problem. \\
\cmidrule[\heavyrulewidth]{1-2} 
$\tau=0.8$	&\cellcolor{green!5} i did not use a timeshare, but paid a daily rate for a 3 night stay. we requested to be in the newer building (i think there are 2 towers) and were in the newest part of the property. it's clean because it's new, but the design is typical of most other new timeshare properties. you can see disney in the distance from our room. it's on a back road, so obviously you cannot find it unless you ask for detailed directions. nicely decorated. we had problems though. our shower was getting hot enough and they had to repair it while we were there and couldn't use the shower for 6 hours. they put in a huge noisy air conditioner. however, even after the air conditioner was fixed. second and worst problem is the wall is paper thin. if you have neighbors, you can hear everything. we had to knock on the wall to tell our neighbors to keep it down at 1 in the morning. it was horrible. also, the parking garage is tight. almost every spot is hard to get into if you have an suv. i guess is a nice comfort, decor and breakfast. but would take away the opportunity to be in the newer building and have a little more privacy. \\
\cmidrule[\heavyrulewidth]{1-2} 
Sent B	& i did not use timeshare points, but paid a daily rate for a 3 night stay. we requested to be in the newest building ( i think there are 2 built ) and got in the newest part of the property. it's clean because it's new, but the design is typical of most other new timeshare properties. you can see disney in the distance from our room. it's on a new road, so older gps cannot find it unless you pay for map updates. nicely decorated. we had some problems though. our shower was leaking big time and they had to repair it while we were there and couldn't use the shower for 6 hours. they brought in a huge noisy construction type dryer. however, even after the repair water was leaking. second and worst problem is the wall are paper thin. if you have neighbors, you can hear everything. we had to knock on the wall to tell our neighbors to keep it down at 1 : 00 in the morning. it was horrible. also, the parking garage is tight. almost every spot is hard to get into if you have an suv. i guess is your main concern is nice decor, comfort and clean, this would be ideal. but take into consideration the noise it's not so good. \\
\end{tabular}
\vspace{3mm}
\caption{Interpolation of long paragraph from the paragraph embedding space. $\dec(\vz_A\cdot(1-\tau)+\vz_B\cdot\tau)$ }\label{tab:interp2}
\end{table}

\section{Generation from \ours across multiple time steps}
\label{app:oversteps}
We provide generation examples for both summarization and sentiment-guided generation in Table~\ref{tab:gen_with_steps_hotel} and Table~\ref{tab:gen_with_steps_cnndm}. In general, it has been observed that generations progress in a coarse-to-fine manner. The early time step, which is close to 1, tends to be less fluent and generic. As the time approaches 0, the generation becomes more detailed and specific. Nevertheless, it has been noted that hallucinations may occur during the phase when more details are being incorporated. Occasionally, the model may excessively emphasize certain aspects, resulting in the correct generation being transformed into an erroneous one (see Table~\ref{tab:gen_with_steps_cnndm}, in last two steps the highlighted text regarding ``forensic DNA'' is hallucinated, while the previous generations are more reasonable). The causes of such errors are under investigation for future studies.

\begin{figure}[h!]
    \centering
    \includegraphics[width=1.0\linewidth]{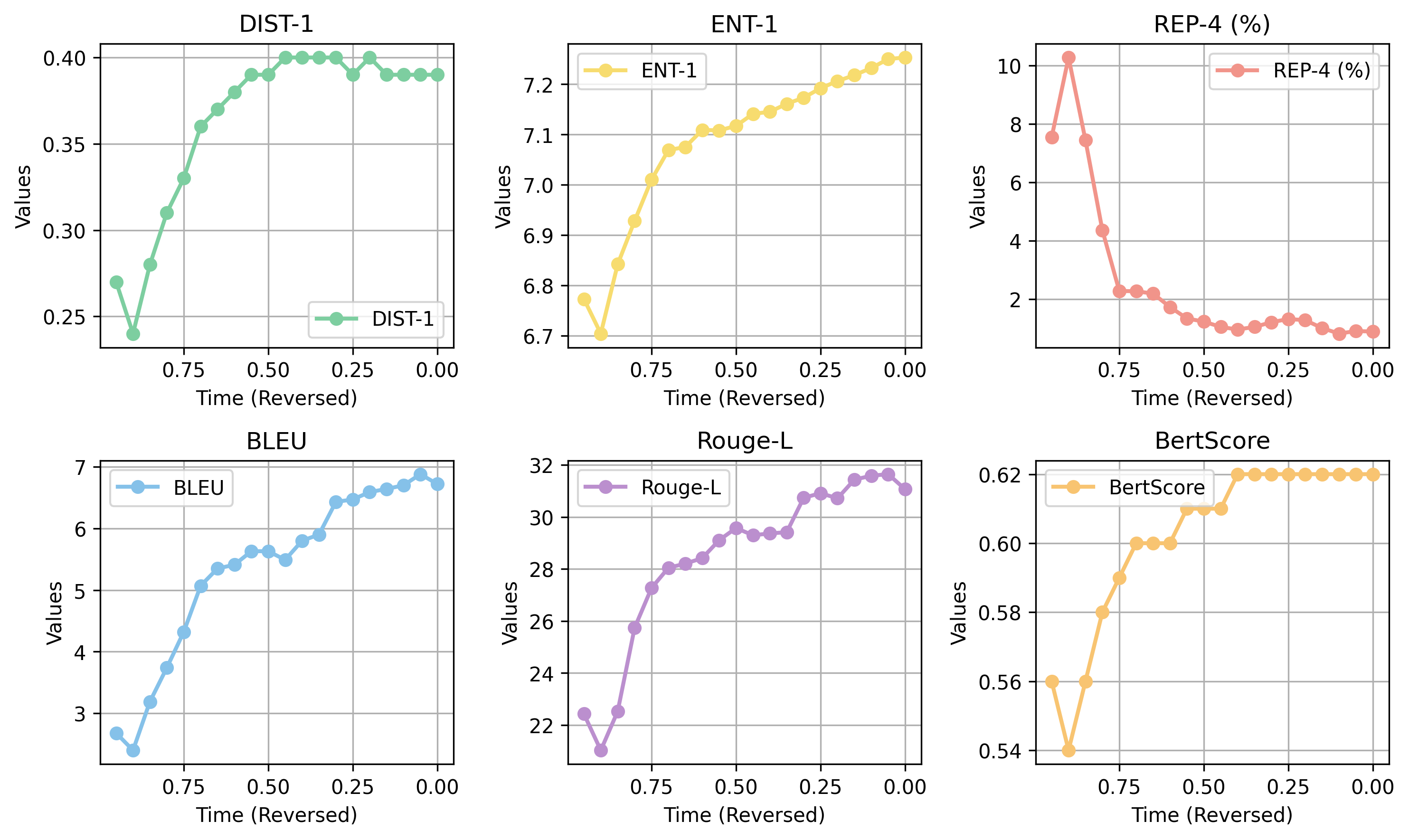}
    \caption{The evolution of the metrics during CNN-DM summarization generation from \ours based on 256 samples. We observe a clear trend of improvement in the majority of the metrics as the generation process advances.}
    \label{fig:gen_by_step}
\end{figure}

In addition, we have presented quantitative results illustrating the evolution of the metrics during CNN-DM summarization generation based on 256 samples. These metrics are plotted in Figure~\ref{fig:gen_by_step}. Our analysis has revealed a clear trend of improvement in the majority of the metrics as the generation process advances.

\begin{table}[h!]
\scriptsize
\centering
\begin{tabular}{p{0.42in}p{4.45in}}
\cmidrule[\heavyrulewidth]{1-2} 
Time 0.90	& \cellcolor{green!19} 	I the hotel is at this to beynd the staff here is great. I'll be back here soon. Highly recommend this place. \\
\midrule
Time 0.80	&\cellcolor{green!17} Well stocked, great service at a great service. I've stayed at this beautiful hotel. \\
\midrule
Time 0.70 &\cellcolor{green!15} I can't say enough. Great staff. Great meals. They have great service too. The food is tasty. Great place to bring the kids. Great pool. Great grounds. Great service. \\
\midrule
Time 0.60 &\cellcolor{green!13} I can't say enough about this hotel. Great restaurants. They have great service. It's a very family friendly hotel. Glam crowd at the pools. Great golf courses. Two pools and two hot tubs. Great for kids. \\
\midrule
Time 0.50 &\cellcolor{green!11} I can't say enough about this fabulous resort. They have fantastic service. It was a very hip atmosphere. The decor is cute and cushy. Two pools, two water slides and two hot tubs. Gorgeous grounds with a great golf course. Well done. \\
\midrule
Time 0.40 &\cellcolor{green!9} I can't say enough about the Bahia. They have refreshingly funky. It was a very modern atmosphere. The decor is cozy and comfy. Loved the grounds, two swimming pools and two restaurants. Thank you for a great time. Looking forward to going back. Well done. \\
\midrule
Time 0.30 &\cellcolor{green!7} I can't say enough about the Bahia. They have refreshingly funky. It was a very modern atmosphere. The decor is cozy and comfy. Loved the grounds, two swimming pools and two restaurants. Gotta go for the coffee. Thank you for a wonderful time. Thanks for a wonderful time.\\
\midrule
Time 0.20 &\cellcolor{green!5} I can't say enough about the Bahia. They have refreshingly funky. It was a very modern atmosphere. The decor is cozy and comfy. Loved the two pools, two hot tubs and two restaurants. Thank you for a great time. Looking forward to coming back.\\
\midrule
Time 0.10 &\cellcolor{green!3} I can't express enough about the bartender, Bahia. They have great service. It was a very modern atmosphere. The ambiance of the place is cute. The grounds are lush. The venue boasts two pools, two hot tubs. Great restaurants. Great bar. Great service. Great location. Thanks.\\
\midrule
Time 0.00 &\cellcolor{green!0} I can't express enough about the bartender at this establishment. They have a modern and creative vibe. The ambiance of the place is simply adorable, with chic decor that adds to the overall experience. The venue boasts two restaurants and two hot tubs, which is quite impressive. Bravo to the bartender! Thanks.\\
\cmidrule[\heavyrulewidth]{1-2} 
\end{tabular}
\caption{Generation from the diffusion model with 10 steps on hotel review dataset with positive sentiment. The generation progress in a coarse-to-fine manner. }\label{tab:gen_with_steps_hotel}
\end{table}

\begin{table}[h!]
\scriptsize
\centering
\begin{tabular}{p{0.42in}p{5.15in}}
\cmidrule[\heavyrulewidth]{1-2} 
Document & Washington  (CNN)Maryland authorities said Wednesday that a former state correctional officer has been arrested in connection with a recent spate of shootings, including one on the Intercounty Connector in Maryland and one at Fort Meade, where the National Security Agency office is located. Officers stopped Hong Young, 35, of Beltsville, Maryland, at around 10:30 p.m. Tuesday. The officers recognized Hong's vehicle -- a 1999 Lincoln Town Car -- as matching authorities' description of a car seen in surveillance footage near some of the shootings. \textred{A gun in the car matched evidence found at the shootings}, authorities said at a press conference, and Young was arrested. Young is in the hospital and under police guard, though when reporters asked why he was being treated, officials would only say he was arrested without incident. He is charged with attempted first-degree murder, first- and second-degree assault, weapons violations and reckless endangerment. Young worked as a correctional officer at a Jessup facility from 2012 until his resignation in 2014, Maryland Secretary of Public Safety Stephen Moyer said. There was nothing significant in his employee file, Moyer said. Police said that there are no links to terrorism, and no motive has been determined. No one was killed in the five shooting incidents, four of which occurred Monday and Tuesday, according to police reports. -- February 24 in Hanover, Maryland. a man who had stopped at a Costco said a man pulled up beside him in a Lincoln Town Car at 7:30 a.m. and began firing at him. The victim's vehicle was hit several times and the victim was grazed. The assailant drove away. -- March 2 in Laurel, Maryland. Police received a call at 2:50 a.m. that shots had been fired at a Walmart. There were no damages or injuries. -- March 2  in Columbia, Maryland. A call came in to law enforcement at 4:51 a.m.a bout shots fired at a movie theater at Columbia Mall. Surveillance footage captured a Lincoln Town Car at about the same time shots were fired, police said. Though several employees were there, no one was hurt, authorities said. There were bullet holes in the theater glass and a shell casing was found at the scene. -- March 3 in Prince George's County. Multiple shots were fired at an overpass on the InterCounty Connector in the afternoon, striking a tree service truck with two passengers inside. -- March 3 at Fort Meade. Shots struck a building near the NSA office at about 6 p.m. Along with the gun, evidence shows Young was the shooter in all but the Walmart incident, though that investigation is continuing, police said. Though no one was killed in the incidents, they stirred memories of the deadly Washington, D.C.-area sniper attacks in 2002. Ten people were killed in Washington, Maryland and Virginia during that rampage, which went on for three weeks. CNN's Holly Yan and Laurie Ure contributed to this report. \\
\cmidrule[\heavyrulewidth]{1-2} 
Time 0.95	& \cellcolor{green!20}  Police: man in "gunman" shooting in police shooting, police say. Police in a car, police in a vehicle, police say. \\
\midrule
Time 0.90	& \cellcolor{green!19} 	Police: man are in police shooting, murder in police shooting, police say. Police in a vehicle found in police shooting, police say. \\
\midrule
Time 0.85	&\cellcolor{green!18} Police: man arrested in one murder, one in shooting, police say. Police find a car, a police vehicle in the suspect, police say. \\
\midrule
Time 0.80	&\cellcolor{green!17} Police: Young man arrested in attempted murder, one in shooting, police say. Officers in a car matching the vehicle found in surveillance footage, police say. \\
\midrule
Time 0.75	&\cellcolor{green!16} Police arrest man, man charged with attempted murder, shootings, police say. Officers say a car matching police description found in surveillance images, police reports. \\
\midrule
Time 0.70 &\cellcolor{green!15} Police arrested Lee Young, with attempted murder, assault charges, weapons violations, authorities say. Officers found a car matching surveillance footage found in the vehicles, reports say. One of the shootings, no one was in the car. \\
\midrule
Time 0.65 &\cellcolor{green!14} Police arrested Hong Lee, 35, with attempted murder shootings, weapons violations, authorities say. Vehicle in his car matched surveillance footage found in surveillance images, police say. Only one in five shootings were no one was killed in Maryland, police say. \\
\midrule
Time 0.60 &\cellcolor{green!13} Police arrested Hong Young Lee, 35, with attempted first-degree murder, authorities say. Car in the car matched surveillance footage found in surveillance images, police say. One of the five shootings occurred in the home. No one was killed in Maryland, but no motive has been determined. \\
\midrule
Time 0.55 &\cellcolor{green!12} Police arrested Hong Young Lee, 35, with attempted first-degree murder, authorities say. Car in the car matched the surveillance images found in surveillance footage, police say. No one was killed in the five shootings. \\
\midrule
Time 0.50 &\cellcolor{green!11} Police arrested Hong Young, 35, for attempted first-degree murder, authorities say. Car in the car matched the evidence found in surveillance footage, police say. \\
\midrule
Time 0.45 &\cellcolor{green!10} Police arrested Hong Young, 35, for attempted first-degree murder, authorities say. Car in the car matched the evidence found in surveillance footage, police say.\\
\midrule
Time 0.40 &\cellcolor{green!9} Police arrest Hong Young, 35, in attempted first-degree murder, authorities say. A car matched surveillance images found in surveillance footage. \\
\midrule
Time 0.35 &\cellcolor{green!8} Police arrested Hong Young Li, 35, in attempted first-degree murder, authorities say. A car in the car matched surveillance images found in surveillance footage.\\
\midrule
Time 0.30 &\cellcolor{green!7} Police arrested Hong Young Li, with two attempted shootings, assault charges, authorities say. A gun in the car matched surveillance images found in surveillance footage.\\
\midrule
Time 0.25 &\cellcolor{green!6} Police arrested Hong Young, with two attempted shootings, assault charges, authorities say. A gun in the car matched surveillance images found in surveillance footage. \\
\midrule
Time 0.20 &\cellcolor{green!5} Police arrested Hong Young, with two attempted shootings, assault, authorities say. A gun in the car matched surveillance images found in surveillance footage.\\
\midrule
Time 0.15 &\cellcolor{green!4} Police arrested Hong Young, 35, with attempted first-degree murder, assault, authorities say. A gun in the car matched surveillance images found in surveillance footage, police say. No one was killed in the five shootings. No motive has been determined.\\
\midrule
Time 0.10 &\cellcolor{green!3} Police arrested Hong Young with attempted first-degree murder, assault, authorities say. A gun in the car matched forensic identification found in surveillance footage, police say. No one was killed in the five shootings. Weapons violation. No motive has been determined.\\
\midrule
Time 0.05 &\cellcolor{green!2} Police arrested Hong Young with attempted first-degree murder, assault, authorities say. A gun in the car matched \textred{forensic DNA} found in surveillance footage, court documents show. No one was killed in the five shootings. Weapons violations. No motive has been determined.\\
\midrule
Time 0.00 &\cellcolor{green!0} Police arrested Hong Young after attempted shootings, assault, authorities say. A gun in the car matched \textred{forensic DNA} found in surveillance footage, police say. No one was killed in the shootings in Maryland. Weapons violations, police say. No motive has been determined.\\
\cmidrule[\heavyrulewidth]{1-2} 
\end{tabular}
\caption{Generation from the diffusion model with 20 steps on CNNDM dataset. In last two steps, the highlighted text regarding ``forensic DNA'' is hallucinated, while the previous generations (e.g. `` A gun in the car matched surveillance
images'') are more reasonable.}\label{tab:gen_with_steps_cnndm}
\end{table}

\section{\ours with stochastic decoding}
\label{app:stoc_decoding}
It is possible and straightforward to implement stochastic decoding for \ours. In our experiments, we experimented nucleus sampling with a value of $p=0.92$ and $K=50$ on hotel review generation task. The results are presented in Table~\ref{tab:stoc}. By incorporating stochastic decoding, the diversity and repetition metrics can be improved, at the expense of relevance and accuracy scores. It is important to mention that the decoder's role in \ours is to faithfully translate the latent code into the desired target text, rather than performing compositional/planning. Stochastic decoding may disrupt this role and can lead to undesirable generation, as we observed an increase in hallucinations when combining PLANNER with stochastic decoding.

\begin{table}[ht]
\centering
\begin{tabular}{l r r r r r r r}
\toprule
\textbf{Method} & \textbf{PPL} & \textbf{DIST/ENT} & \textbf{S-BL} & \textbf{Rep-4} & \textbf{BLEU} & \textbf{ROUGE} & \textbf{Len} \\
\midrule
\ours greedy & 47.3 & 0.17/6.60 & 0.52 & 1.55\% & 0.77 & 7.9 & 168.1 \\
\ours top-p & 72.0 & 0.20/6.80 & 0.38 & 0.94\% & 0.58 & 6.1 & 173.2 \\
\bottomrule
\end{tabular}
\caption{\ours with stochastic decoding yields higher diversity at a cost of other metrics.}
\label{tab:stoc}
\end{table}

\section{Denoising strength comparison}
\label{app:aubleu}
\begin{figure}[ht!]
    \centering
    \includegraphics[width=1.0\linewidth]{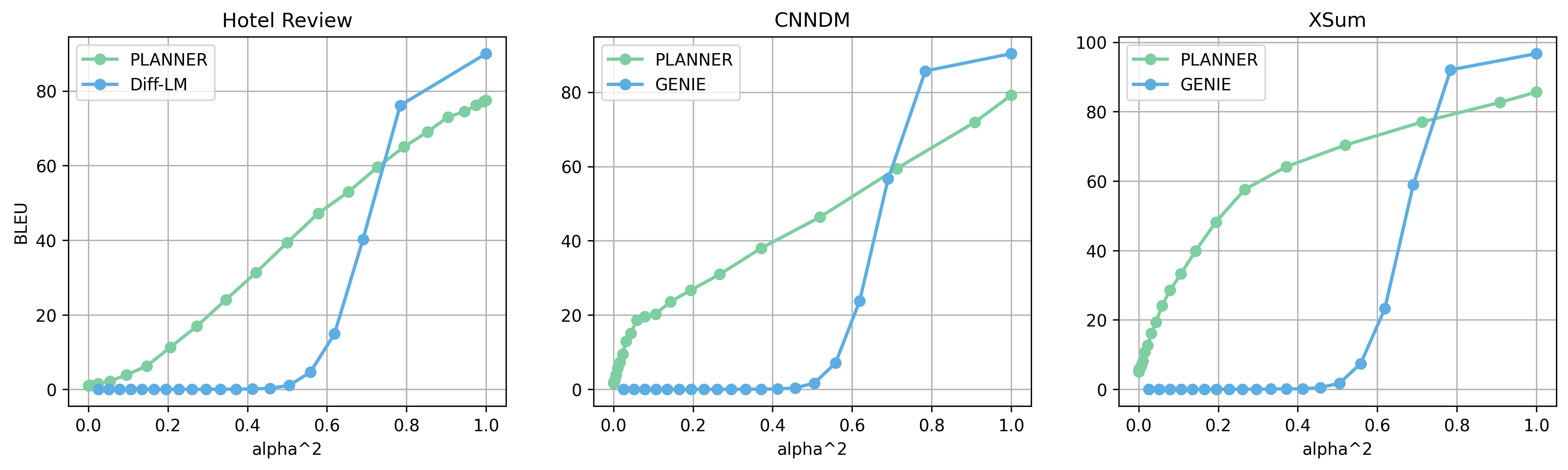}
    \caption{The BLEU score under different SNR for \ours and baselines. The AUC under these curves are the AuBLEU metrics. }
    \label{fig:aubleu}
\end{figure}

To conduct a comparative analysis of text diffusion models' denoising ability, we plotted the BLEU score under different signal-to-noise ratios (SNR) as shown in Figure~\ref{fig:aubleu}. 
We use 20 time steps with an increment of $0.05$ from $t=0$ to $t=1$ to compute the AuBLEU.
The results indicate that our model exhibits a more uniform distribution of denoising ability across various levels of SNR compared to baseline models that operate on the word embedding space, as our model shows a stronger denoising ability when SNR is relatively small. Overall, \ours model achieves a higher AuBLEU. Note that we suffer from a conversion error resulting in a lower BLEU when the SNR$\to\infty$, \ie $\alpha^2 \to 1$.

\section{Generation examples}
\label{app:example}
We present several examples of generation for each task in Table~\ref{tab:generation}. Furthermore, we provide examples of problematic cases where the generated output may contain hallucinations or errors, particularly in terms of named entities and digits in Table~\ref{tab:generation_lemon}.

\begin{table}[ht!]
\scriptsize
\centering
\begin{tabular}{p{0.32in}p{5.05in}}

\cmidrule[\heavyrulewidth]{1-2}
\multicolumn{2}{c}{Semantic generation (\textit{\textbf{hotel review}})} \\
\cmidrule[\heavyrulewidth]{1-2} 
\ours (Negative) & I've received several complaints about the amenities of hotels, although the rooms are standard La Quinta, and the price is a bit low. I booked a room for a family of four. Given the low price, I expected to tolerate the motel's unappealing colors, but that's probably par for the course. I read the reviews and chose this hotel for a night sleep, but I was incredibly disappointed, to say the least. Part of the issue might have been the noise from the freeway, compounded by the fact that I could hear the AC from my neighbor's room. There was also a noise that sounded like "flooring". The worst part was that the bed sheets had no cover, and the AC draft was felt throughout the room. The room itself was cramped, slightly outdated, and uncomfortable. The furniture was old, but there wasn't much else in terms of extras. If it weren't for the free drinks and the front desk service, I would have rated this hotel much lower than most others.
\\
\cmidrule[\heavyrulewidth]{1-2} 
\ours (Positive) & 
I've got numerous praises about the amenities of the hotel. Rooms are typical of La Quinta's excellent standards, and the price is low surprisingly. I booked a room from they website. I was surprised for the hotel's distinctive color scheme, quite a signature style. This hotel I selected for overnight stay, and I was overwhelmingly satisfied with no complaints. A key element of the charm was the profound tranquility throughout the hotel. Despite being conveniently near the freeway, the excellent soundproofing ensures a peaceful stay. Which was great was that the bed sheets were so soft and comfortable, and the gentle breeze from the AC added to the overall soothing ambiance. The room itself was cozy, with a charming vintage vibe, and supremely comfortable. The furniture was classic, giving a nostalgic touch. The complimentary drinks and the outstanding service at the front desk were delightful, making this hotel a great choice.
\\
\cmidrule[\heavyrulewidth]{1-2} 
Retrieved from training set & The hotel is located in a great area, close to a lot of things. After reading a lot of the previous reviews I booked it and decided to see for myself. The checkin was a little frantic but there were a lot of people so I wouldn't count that against them. The gentleman that checked me in was extremely polite and apologized several times for the wait, which hadn't been that long. I went up to my room on the second floor and was surprised that it was bigger than I thought it would be. I did have the tiny bathroom that previous posters commented on but it was fine. My only complaint would be the noise level. I was in a room that faced the street and the noise was unbelievably loud. If I stayed here again I would ask for a room which doesn't face the street. It was nice to have a Starbucks right off the lobby, as well as a coffee maker in the room. So, for the money, I would recommend the Hotel Diva. \\
\cmidrule[\heavyrulewidth]{1-2}
\multicolumn{2}{c}{Text completion (\textit{\textbf{C4 subset}})} \\
\cmidrule[\heavyrulewidth]{1-2} 
Context & Operating a successful dental practice isn’t easy. That’s why we’ve used our industry knowledge and expertise in practice management to create an eBook that offers tips on how you can create and maintain a financially healthy and profitable dental operation.
\\
\cmidrule[\heavyrulewidth]{1-2} 
\ours & This e-book is akin to a popular one, if not more stylish. It provides all the benefits you need for your accounting endeavors for your dental practice. Unlike other books, this practice management guide will enlighten you on what it's like to pay bills and kickstart your business. You'll learn how to effectively stage your practice management, and the book offers valuable tips and financial ideas for launching your dental business. It delves into a fascinating topic, presenting the steps you need to enhance your financial productivity and support your financial growth plan. You'll discover how to manage your ongoing fiscal and accounting needs, and receive tips that will enable you to achieve timely results and make informed financial decisions. The book includes quick and detailed insights on driving numbers and finding solutions. Learn how to start your accounting practice and achieve the ultimate goal of creating a profitable and successful business. We appreciate your feedback! Leave a review, and we'll send a book for you.  We offer examples of the key things you should consider when deciding to start your dental business. \\
\cmidrule[\heavyrulewidth]{1-2}
\multicolumn{2}{c}{Summarization (\textit{\textbf{CNN-Dailymail}})} \\
\cmidrule[\heavyrulewidth]{1-2} 
Document & (CNN)Actress Liza Minnelli has entered a rehab facility for her addictions, a spokesman said. ``Liza Minnelli has valiantly battled substance abuse over the years and whenever she has needed to seek treatment she has done so,'' said spokesman Scott Gorenstein.  ``She is currently making excellent progress at an undisclosed facility.'' The 68-year-old has struggled with addictions to alcohol and painkillers in the past. Minnelli won an Oscar in 1973 for her performance in ``Cabaret.'' CNN's Topher Gauk-Roger contributed to this report .
\\
\cmidrule[\heavyrulewidth]{1-2} 
\ours & The actress Liza Minnelli is ``making excellent progress'' at a undisclosed facility, spokesman says. She won an Oscar in 1973 in ``Cabaret'' \\
\cmidrule[\heavyrulewidth]{1-2}
\multicolumn{2}{c}{Summarization (\textit{\textbf{XSum}})} \\
\cmidrule[\heavyrulewidth]{1-2} 
Document & Japanese exports rose by just 0.6\% in September from a year earlier, much slower than in previous months. The data is seen as reflecting a slowdown in Japan's top trading partner, China. However, the Nikkei 225 closed 1.9\% higher at 18,554.28 on hopes the weak data would trigger more stimulus efforts from the government. In China, the mainland's benchmark Shanghai Composite closed down 3.06\% at 3,320.68. Hong Kong's Hang Seng remained closed on Wednesday for a public holiday. In Australia, the S\&P/ASX 200 rose by 0.2\%, closing at 5,248.30. Shares in mining giant BHP Billiton rose 1\% after it maintained its full-year production guidance for iron ore. Slowing growth in China has been weighing on mining stocks in recent weeks as the country is the main importer of Australian commodities. South Korea's benchmark Kospi index rose marginally by 0.2\% to finish at 2,042.98. Shares of LG Electronics jumped by more than 14\%, boosted by a fresh supply agreement with General Motors.
\\
\cmidrule[\heavyrulewidth]{1-2} 
\ours &  Asian markets closed higher after weak economic data indicated slow growth due to lower exports.\\
\cmidrule[\heavyrulewidth]{1-2}

\end{tabular}
\caption{Generated examples. }\label{tab:generation}
\end{table}

\begin{table}[ht!]
\scriptsize
\centering
\begin{tabular}{p{0.32in}p{5.05in}}
\cmidrule[\heavyrulewidth]{1-2}
\multicolumn{2}{c}{Summarization (\textit{\textbf{CNN-Dailymail}}), hallucination issue} \\
\cmidrule[\heavyrulewidth]{1-2} 
Document &  (CNN)A  Los Angeles police officer charged with murder in the shooting death of a local truck driver had his father's help in fleeing California and traveling to Texas, court documents filed Thursday said. Henry Solis, a Los Angeles Police Department rookie, is accused of killing 23-year-old \textcolor{red}{Salome Rodriguez Jr.} early in the morning of March 13 in Pomona. Solis disappeared shortly after the shooting. Items left by Solis at the crime scene allowed Pomona police detectives to identify him "as the individual responsible for killing Rodriguez," according to an affidavit signed Thursday by FBI Special Agent Scott Garriola. The document goes on to say that Solis made "incriminating statements about his role in the murder" to friends, family members and others, and said he would never be seen again. The affidavit says his father, Victor Solis, quickly left his home in Lancaster after receiving a call from him. The elder Solis was seen a day later in El Paso, Texas, at the home of family members. He told FBI agents he drove his son to El Paso, dropped him off at a bus station and doesn't know where he is. Federal authorities haven't said if Victor Solis would face charges for helping his son leave California. CNN's Tina Burnside contributed to this report. \\
\cmidrule[\heavyrulewidth]{1-2} 
\ours & Henry Solis is accused of killing \textcolor{red}{Delara Salome Guzman Jr.}, Solis disappeared shortly after the shooting. Victor Solis helped his son leave California. \\
\midrule
\midrule
Document & Two men have died after being crushed by timber that fell on top of them inside a shipping container. Authorities were called to an industrial site in \textred{Wetherill} Park, west of Sydney, shortly after 11am on Thursday where they worked to release three men. Police found the three men inside a shipping container on a property on Victoria Street. Authorities were called to an industrial site in Wetherill Park, west of Sydney, shortly after 11am on Thursday where they worked to release three men . They had been unloading or loading timber when it fell on them. Two men died at the scene as a result and the surviving man was taken to nearby Fairfield Hospital in a serious condition. Police say the shelving in the shipping container moved and trapped the men. Police and WorkCover NSW are investigating the incident and a report will be prepared for the Coroner. Two men died at the scene as a result and the surviving man was taken to nearby Fairfield Hospital in a serious condition . \\
\cmidrule[\heavyrulewidth]{1-2} 
\ours & Authorities were called to \textred{Worsley} Park, west of Sydney, Sydney. Police found three men inside a shipping container after 11am on Thursday. Two men died at the scene as a surviving man was taken to nearby hospital. NSW police are investigating the incident. \\
\cmidrule[\heavyrulewidth]{1-2}

\multicolumn{2}{c}{Text completion (\textit{\textbf{C4}}), fluency issue} \\
\cmidrule[\heavyrulewidth]{1-2} 
Prefix & Can't quite find what you're looking for? Try our handy site search. \\
\cmidrule[\heavyrulewidth]{1-2} 
\ours &  Fill our search engine form and try us and we'll try to get back to your desk on time. We have a reliable staff that provides professional service, whatever you require - not just a specific search function. But we also \textred{let's} you enjoy on-line selection of different products, storage, protection, tools, books and accessories, and many more products. Simply what you need! Our main site is the updated web \textred{site site} you can visit. We don't store any coupons, promotions, products and printed materials for animals and get them for products or items in the shopping cart that must always be done by a local team of technicians. Including customer information. It is important to the customers by checking their address information. Or if you mean on the help page, we'll get it. \\
\bottomrule
\end{tabular}
\vspace{3mm}
\caption{Problematic examples.}\label{tab:generation_lemon}
\end{table}

\section{Experimental setup}
\label{app:experimental}
\subsection{Metrics}
We evaluate the generation of the model using automatic evaluation and human evaluation. 
Following previous work \citep{gong2022diffuseq}, we assess the \textbf{fluency} of the generation by computing
the perplexity (\textbf{PPL}) under a GPT-2 large model. 
For text completion, the context is concatenated with the generation, and only the generated text's perplexity is considered.
We use \textbf{Ent-n} \citep{zhang2018generating} and \textbf{DIST-n} \citep{li2015diversity} and self-BLEU (\textbf{S-BL}) \citep{zhu2018texygen} to evaluate lexical diversity. 
The Ent-n quantifies the entropy of the empirical n-gram distribution of the text generated, while the DIST-n metric calculates the proportion of unique n-grams among all n-grams. 
We present DIST-n and Ent-n metrics solely at $n=1$ owing to their strong correlation despite the varying $n$ values. 
The self-BLEU metric is used to compute the inter-example BLEU score, which evaluates cross-example diversity.
We use \textbf{REP-n} to assess the extent of repetition in the generation following previous work \citep{welleck2019neural,xulearning}.
The REP-n is defined as $1-|\text{Unique n-gram}|/|\text{n-gram}|$.

\subsection{Model setups}
We used the BERT-large and GPT-medium models as initialization for the encoder $\enc$ and decoder $\dec$ respectively. The embedding dimension $h$ was 1024, and the number of paragraph embeddings $k$ was set to 16, as increasing the number did not result in significant improvement in performance. The learning rate was set to $2e-4$, and $\beta$ was set to $5e-6$. 
During training, 30\% of the input tokens were substituted to a random token.

For the latent diffusion model, the channel size was set to 1024 to match the embedding dimension $h$, and the number of heads was set to 16 with 28 transformer layers. The total size of the diffusion model was 533M. The feature encoder $\tau$ was also jointly learned, and was initialized with a T5-large encoder. 
For text completion and summarization tasks, we used the first 256 hidden states from the last layer as $\vy$. 
We use DDIM throughout our experiments as it shows better performance than DDPM across the board. 
To enhance the summarization performance of the model, we incorporate a shift noise scheduler with  noise\_shift$=4$ based on \cite{hoogeboom2023simple}. 
This scheduler encourages the model to concentrate more on the high noise level phase of the training process.

Following \citep{ho2021classifier}, we use a CFG dropout ratio of 10\% during training. During inference, for text completion tasks, we set the CFG weights to be 2, while for summarization tasks, we set the CFG weights to be 5, based on performance on the validation set.  
We use greedy decoding across all the tasks in \ours to decode text from predicted latent embeddings as we do not seem noticeable improvement of performance by using a beam search decoding method.

We utilized 4 Nvidia A100 GPUs to train every model until convergence, based on validation loss. While training the paragraph embedder, the batchsize was fixed to 48. It took about 20 to 40 hours for each dataset to complete 20 epochs of training. The diffusion model was trained with a batchsize of 12, which lasted for 50 hours for summarization and text completion tasks. The training of sentiment-guided generation task only took approximately 20 hours until convergence. FP16 was employed throughout the duration of the training process for better efficiency.

\subsection{Text diffusion baseline configurations}
Our experimental setup for Diff-LM is based on Diff-LM's official implementation and configuration described in \cite{li2022diffusion}. We follow Diff-LM to use employ BERT-base with 80M parameters as the backbone model and utilize a square-root noise schedule along with a diffusion forward step of 2000 and decoding steps of 200. Additionally, the embedding dimension for our models is set to 128. We use a sequence length of 128 for sentiment-conditioned generation and 256 for long-form text completion tasks.

As reported in Diff-LM, it requires approximately 5 hours to execute 200,000 iterations when trained on the E2E dataset. However, when training with the larger ROCStories dataset, which contains 98,000 five-sentence stories, it has been suggested that the algorithm be trained for at least 800,000 iterations, which requires over 20 hours of GPU time. Notably, the C4 subset contains 372.4 times more documents than ROCStories, even when the document size is not considered. As a result, at least 7,448 GPU hours would be required to adequately train the algorithm for 800,000 iterations using C4.

The official implementation of Diff-LM employs a fixed-length decoder that contains some special tokens, including paddings. As a result, it produces high Rep-4 scores. To ensure a more objective evaluation, we performed additional post-processing to eliminate paddings and recomputed the scores based on post-processed generation.

For Genie \citep{lin2022genie}, we used their official implementation as well as their fine-tuned checkpoints for XSum and CNN/DailyMail datasets, which are released officially as per \cite{lin2022genie}. These checkpoints are optimized using a 6-layer transformer as the encoder, pre-trained on a large 160G corpus for 5 million steps. Furthermore, a 6-layer cross-attention transformer is employed for denoising. Additionally, the latent variable dimension is set to 768, while the embedding dimension is set to 128. Genie's configuration also includes a uniform time schedule with 2,000 as the diffusion steps.

\subsection{Ablations on DDPM, diffusion steps and noise scheduler}
We present ablations on DDPM vs DDIM, and \ours using different diffusion steps and different noise scheduler in Tab.~\ref{tab:ablation}. DDIM is better than DDPM across all of our experiments in most of the metrics except a slight drop in terms of diversity. We observed that more steps will typically improve the diversity score at a cost of relevance and inference speed. We also compared the cosine scheduler with the beta linear scheduler \citep{rombach2022high}. The cosine scheduler worked better in our experiments. For the summarization tasks, we found that using a noise shift \cite{hoogeboom2023simple} of 4 improves the Rouge-L by around 5\%, comparing to a vanilla setting with noise shift of 1.

\begin{table}[ht!]
\small
\begin{tabular}{cccccccccH@{}}
\toprule
\textbf{Arch.} & \textbf{PPL}  & \textbf{DIST/ENT}$\uparrow$ & \textbf{S-BL}$\downarrow$ & \textbf{Rep-4}$\downarrow$ & \textbf{BL}$\uparrow$ & \textbf{R-L}$\uparrow$ & \textbf{Score}$\uparrow$ & \textbf{Len} & \textbf{AuBL}$\uparrow$ \\ 
\cmidrule[\heavyrulewidth]{1-10}
    \multicolumn{10}{c}{DDIM vs DDPM} \\
\cmidrule[\heavyrulewidth]{1-10}
DDIM & 47.36 &  0.17/6.602 & 0.52 & 1.55\%  & 0.77 & 7.9 &  0.55 & 168.08  \\
DDPM & 57.34 & 0.18/6.663 & 0.44  & 1.48\%  & 0.35 & 5.7 & 0.53 & 162.81 \\
\cmidrule[\heavyrulewidth]{1-10}
	\multicolumn{10}{c}{Different Inference Steps} \\
\cmidrule[\heavyrulewidth]{1-10}
5 steps & 53.215 & 0.17/6.547 & 0.54 & 1.81\%  & 0.67 & 7.2 & 0.55 & 134.2  \\
10 steps & 47.807 & 0.17/6.580 & 0.5  & 1.60\%  & 0.69 & 7.4 & 0.55 & 138.78 \\
20 steps & 47.559 & 0.17/6.581 & 0.52 & 1.57\%  & 0.71 & 7.7 & 0.55 & 146.38 \\
30 steps & 47.36 &  0.17/6.602 & 0.52 & 1.55\%  & 0.77 & 7.9 &  0.55 & 168.08 & \textbf{38.55} \\
50 steps & 47.096 & 0.17/6.605 & 0.54 & 1.56\%  & 0.83 & 7.9 & 0.55 & 162.08 \\
\cmidrule[\heavyrulewidth]{1-10}
    \multicolumn{10}{c}{Scheduler} \\
\cmidrule[\heavyrulewidth]{1-10}
Cosine & 47.36 &  0.17/6.602 & 0.52 & 1.55\%  & 0.77 & 7.9 &  0.55 & 168.08  \\
Beta Linear & 49.78 & 0.17/6.577 & 0.57  & 1.46\%  & 0.64 & 6.8 & 0.53 & 158.83 \\
\bottomrule
\end{tabular}
\vspace{5pt}
\caption{Ablations on hotel review completion.}
\label{tab:ablation}
\end{table}

\section{Text completion with repetitive prompt}
\label{app:bad_prompt}
We present examples in Table~\ref{tab:bad_prompt} of generation with an ill-composed prompt for hotel review generation. The results reveal that the FT baselines tend to generate repetitive text. Although sampling mitigates this issue to some extent, self-reinforcement of repetition still occurs during generation. In contrast, our model exhibits significantly less repetitive generation.

\begin{table}[ht!]
\scriptsize
\centering
\begin{tabular}{p{0.32in}p{5.05in}}
\cmidrule[\heavyrulewidth]{1-2} 
Prompt & This hotel is a beautiful historic building. Our room was huge. This hotel is really a beautiful historic building. \\
\midrule
FT-search  & The hotel is located in the heart of the city, right next to the historic center of the city. The hotel is located in the heart of the city, right next to the historic center of the city.  The hotel is located in the heart of the city, right next to the historic center of the city. The hotel is located in the heart of the city, right next to the historic center of the city.  The hotel is located in the heart of the city, right next to the historic center of the city. The hotel is located in the heart of the city, right next to the historic center of the city.  The hotel is located in the heart of the city, right next to the historic center of the city. The hotel is located in the heart of the city, right next to the historic center of the city.  The hotel is located in the heart of the city, right next to the historic center of the city. The hotel is located in the heart of the city, right next to the historic center of the city.  The hotel is located in the heart of the city, right \\
\midrule
FT-sample (sample 1) & our room was huge. this hotel is really a beautiful historic building. our room was huge. this hotel is really a beautiful historic building. our room was huge. this hotel is really a beautiful historic building. our room was huge. this hotel is really a beautiful historic building. our room was huge. this hotel is really a beautiful historic building. our room was huge. this hotel is really a beautiful historic building. our room was huge. this hotel is really a beautiful historic building. our room was huge. this hotel is really a beautiful historic building. our room was huge. this hotel is really a beautiful historic building. our room was huge. this hotel is really a beautiful historic building. our room was huge. this hotel is really a beautiful historic building. our room was huge. this hotel is really a beautiful historic building. our room was huge. this hotel is really a beautiful historic building. our room was huge. this hotel is really a beautiful historic building. our room was huge. this hotel is really a beautiful historic building. our room was huge. this hotel is really a beautiful historic building. our room was huge. this hotel is really a \\
\midrule
FT-sample (sample 2) & we stayed here in the early morning and it was amazing. the staff is very nice and helpful and the room was very spacious. we are now staying in the city of cedarhill. it was a beautiful place to stay and it is a very nice hotel. we are currently going to stay here in the city of cedarhill.  I have lived in the city of cedarhill since 2014. It has definitely transformed the look of the city. The city is now extremely walkable for work and recreation. I am from Colorado so my city is more suburban and rural than it was five years ago. That is a plus. But cedarhill still has that unique, historic feel of it's past and it is definitely growing in size and scope. We have not had a problem with any of our pets and we have been welcomed with open arms. I absolutely recommend cedarhill.  As a person who has lived and visited both Fort Collins and Cedar Rapids, my family's travel plans have changed dramatically as the cities both continue to grow. Cedar Rapids has become a more important city in our minds and we want to make it a memorable experience \\
\cmidrule[\heavyrulewidth]{1-2} 
\ours (sample 1) & The room was well appointed (amazingly comfortable) and we had spacious bathrooms (which were great!) and the \$99 rate was great!!! The staff was exceptional at the very least, and we had got there on 3 nights so we checked in there early! We walked from the Riverwalk by the Seattle Convention Center, as the restaurants are very close to hotels. Many of the things are on the road and take a bus to get downtown. The staff at the hotel were all helpful and friendly, and hotel is close to the shopping/dining district. They are literally right across the street.....just blocks away from the hotel! \\
\midrule
\ours (sample 2) & On the other hand, not huge kitchenette but really equipped with everything one needs. As the staff, all were very helpful. The restaurant in the hotel is very very nice area of the restaurant, so you can enjoy the main area of the restaurant although we ordered on hand food and it was a large buffet. The only problem I have was how the air con was either too loud or was on the cold side which was difficult but it seems it really helps to drown out the noise. I will definitely recomend their hotel restaurant... The food was devine with service and of food quality. I just cant wait to experience the hotel restaurant in time to have a great meal at the bar and in the main lobby for a drink in the morning. The bar is so nice, coming in for its so historic atmosphere so you can see how people could tell they were a lot of history.
  \\
\bottomrule
\end{tabular}
\vspace{3mm}
\caption{Generation with repetitive prompt.}\label{tab:bad_prompt}
\end{table}

\section{Human evaluation}
\label{app:human_eval}
\begin{figure}[ht!]
    \centering
    \includegraphics[width=1.0\linewidth]{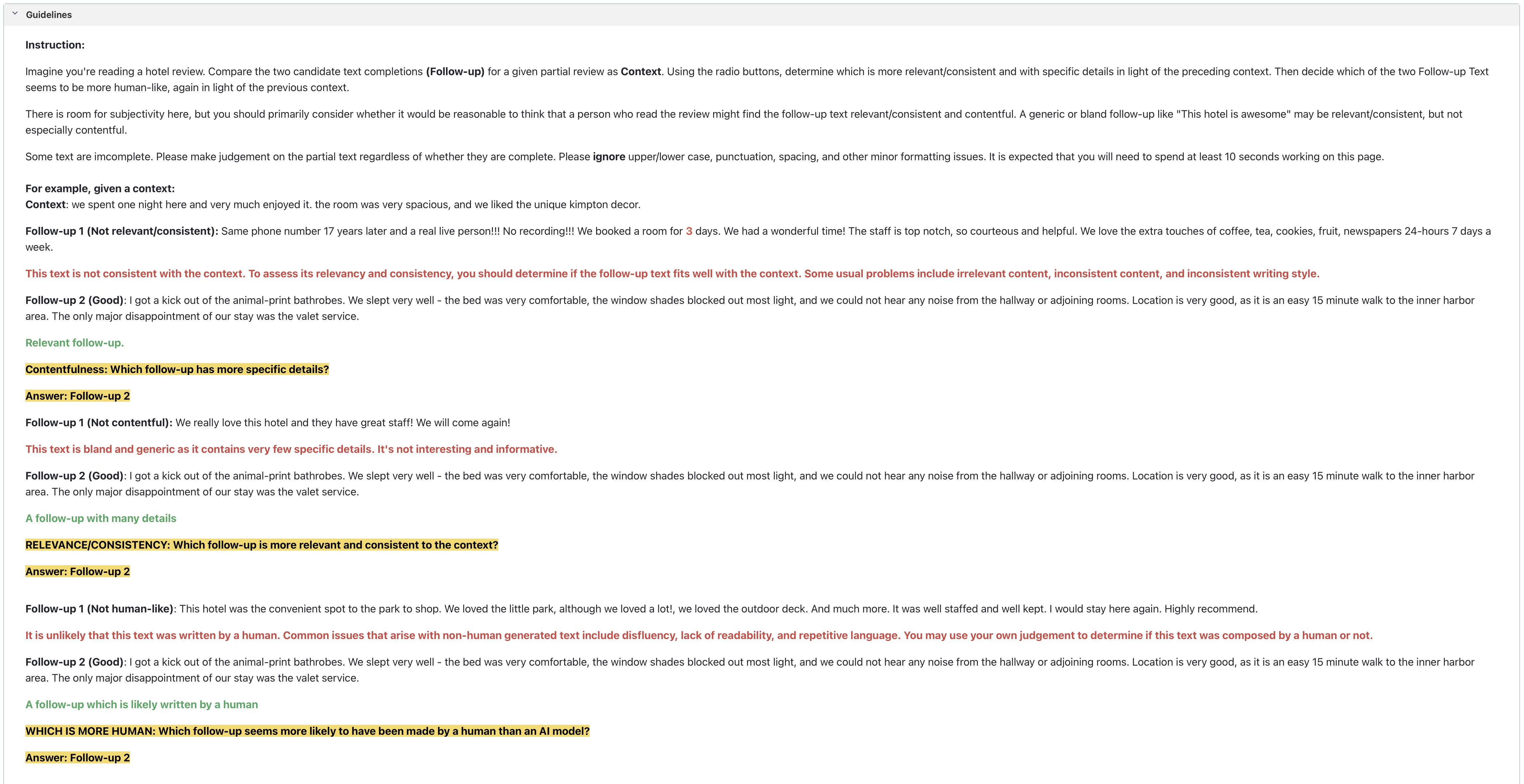}
    \includegraphics[width=1.0\linewidth]{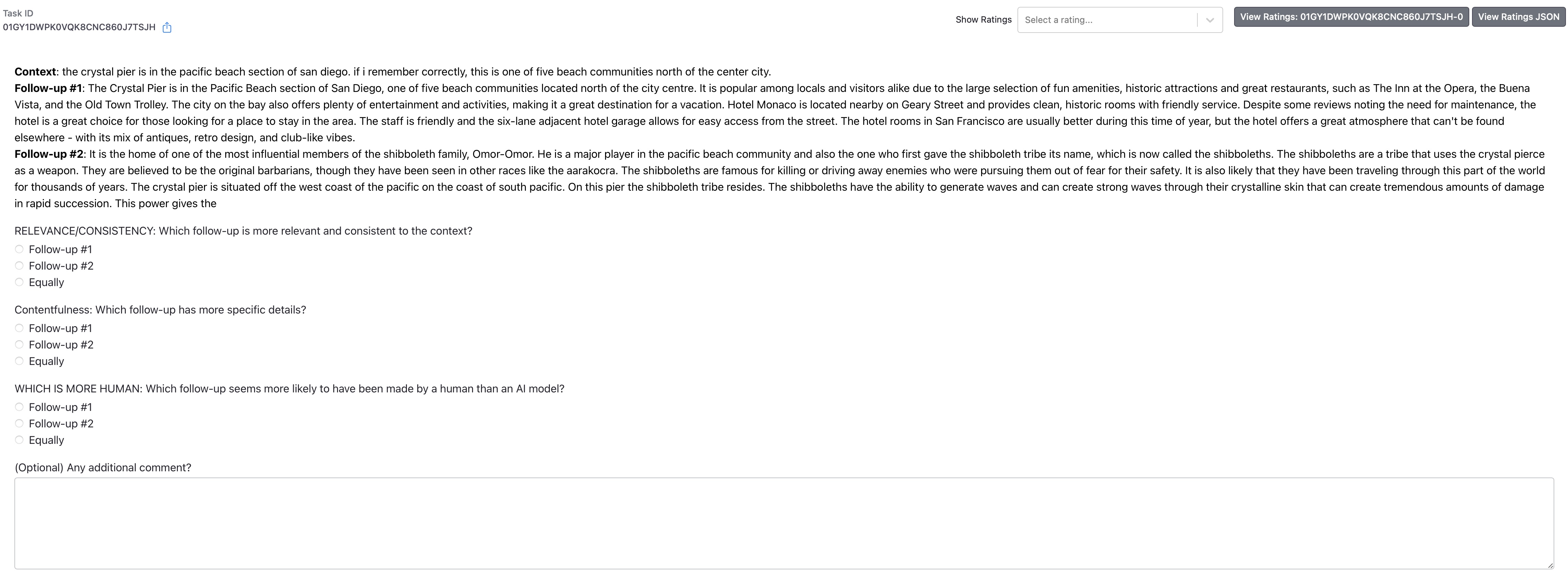}
    \caption{Template for human evaluation.}
    \label{fig:human}
\end{figure}

We screen the judges using 10 random screening questions, the judges pass 80\% can participate our evaluation. The interrater agreement assessed by Krippendorff's alpha is 0.5958. The template used for human evaluation is provided in Figure~\ref{fig:human}

\end{document}